\documentclass[sigconf]{acmart}
\acmSubmissionID{113}

\AtBeginDocument{%
  }

\usepackage{overpic}
\usepackage{amssymb,amsbsy,amsfonts}
\usepackage{color,balance}
\usepackage[fleqn,tbtags]{mathtools}
\usepackage{enumitem}
\usepackage{siunitx}
\usepackage{microtype}

\def \etal {{\emph{et al}.\thinspace}}

\def \eg {{\emph{e.g}.\thinspace}}
\def \ie {{\emph{i.e}.\thinspace}}

\newcommand{\textred}{\textcolor[rgb]{1.0,0.0,0.0}}

\usepackage{multirow}

\copyrightyear{2023} 
\acmYear{2023} 
\setcopyright{acmlicensed}\acmConference[SIGGRAPH '23 Conference Proceedings]{Special Interest Group on Computer Graphics and Interactive Techniques Conference Conference Proceedings}{August 6--10, 2023}{Los Angeles, CA, USA}
\acmBooktitle{Special Interest Group on Computer Graphics and Interactive Techniques Conference Conference Proceedings (SIGGRAPH '23 Conference Proceedings), August 6--10, 2023, Los Angeles, CA, USA}
\acmPrice{15.00}
\acmDOI{10.1145/3588432.3591482}
\acmISBN{979-8-4007-0159-7/23/08}

\definecolor{dblue}{rgb}{0.0,0.0,0.5}
\definecolor{dgreen}{rgb}{0.0,0.5,0.0}
\definecolor{dred}{rgb}{0.6,0.0,0.0}
\definecolor{dorange}{rgb}{0.6,0.25,0.0}
\definecolor{dyellow}{rgb}{0.5,0.5,0.0}

\newcommand{\ignorethis}[1]{}

\newcommand{\intd}{\,\text{d}}

\renewcommand{\textrightarrow}{$\rightarrow$}

\newcommand{\SceneName}[1]{\textsc{#1}}

\newcommand{\C}{C}
\newcommand{\D}{D}

\newcommand{\camera}{\textbf{o}}
\newcommand{\viewDir}{\textbf{v}}
\newcommand{\lightPos}{\textbf{l}}
\newcommand{\pos}{\textbf{p}}
\newcommand{\cues}{\Theta}
\newcommand{\normal}{\textbf{n}}
\newcommand{\f}{f}
\newcommand{\sdf}{\f(\pos)}
\newcommand{\Loss}{\mathcal{L}}
\newcommand{\w}{w}
\newcommand{\s}{s}
\newcommand{\feat}{\bar{\f}}

\begin{document}
\title[Relighting Neural Radiance Fields with Shadow and Highlight Hints]{Relighting Neural Radiance Fields with \\ Shadow and Highlight Hints}

\author{Chong Zeng}
\authornote{Work done during internship at Microsoft Research Asia.}
\affiliation{%
  \institution{State Key Lab of CAD and CG, Zhejiang University}
  \city{Hangzhou}
  \country{China}
}
\orcid{0009-0004-6373-6848}
\email{chongzeng2000@gmail.com}

\author{Guojun Chen}
\affiliation{%
  \institution{Microsoft Research Asia}
  \city{Beijing}
  \country{China}
}
\orcid{0000-0002-3207-6283}
\email{guoch@microsoft.com}

\author{Yue Dong}
\affiliation{%
  \institution{Microsoft Research Asia}
  \city{Beijing}
  \country{China}
}
\orcid{0000-0003-0362-337X}
\email{yuedong@microsoft.com}

\author{Pieter Peers}
\affiliation{%
  \institution{College of William \& Mary}
  \city{Williamsburg}
  \country{USA}
}
\orcid{0000-0001-7621-9808}
\email{ppeers@siggraph.org}

\author{Hongzhi Wu}
\affiliation{%
  \institution{State Key Lab of CAD and CG, Zhejiang University}
  \city{Hangzhou}
  \country{China}
}
\orcid{0000-0002-4404-2275}
\email{hwu@acm.org}

\author{Xin Tong}
\affiliation{%
  \institution{Microsoft Research Asia}
  \city{Beijing}
  \country{China}
}
\orcid{0000-0001-8788-2453}
\email{xtong@microsoft.com}

\renewcommand{\shortauthors}{Zeng et al.}

%!TEX root = ../NeuralRelightHint.tex

\begin{abstract}

  This paper presents a novel neural implicit radiance representation for free
  viewpoint relighting from a small set of unstructured photographs of an
  object lit by a moving point light source different from the view
  position. We express the shape as a signed distance function modeled by a
  multi layer perceptron.  In contrast to prior relightable implicit neural
  representations, we do not disentangle the different light transport
  components, but model both the local and global light transport at each
  point by a second multi layer perceptron that, in addition, to density
  features, the current position, the normal (from the signed distance
  function), view direction, and light position, also takes shadow and
  highlight hints to aid the network in modeling the corresponding high
  frequency light transport effects.  These hints are provided as a
  suggestion, and we leave it up to the network to decide how to incorporate
  these in the final relit result.  We demonstrate and validate our neural
  implicit representation on synthetic and real scenes exhibiting a wide
  variety of shapes, material properties, and global illumination light
  transport.
\end{abstract}



%
% End generated code
%


%!TEX root = ../NeuralRelightHint.tex
\newcommand{\teaserImgWidth}{0.168\textwidth}

\begin{teaserfigure}
\renewcommand{\arraystretch}{0.3}
\addtolength{\tabcolsep}{-5.5pt}
\begin{tabular}{ cccccc }
  \includegraphics[height=\teaserImgWidth]{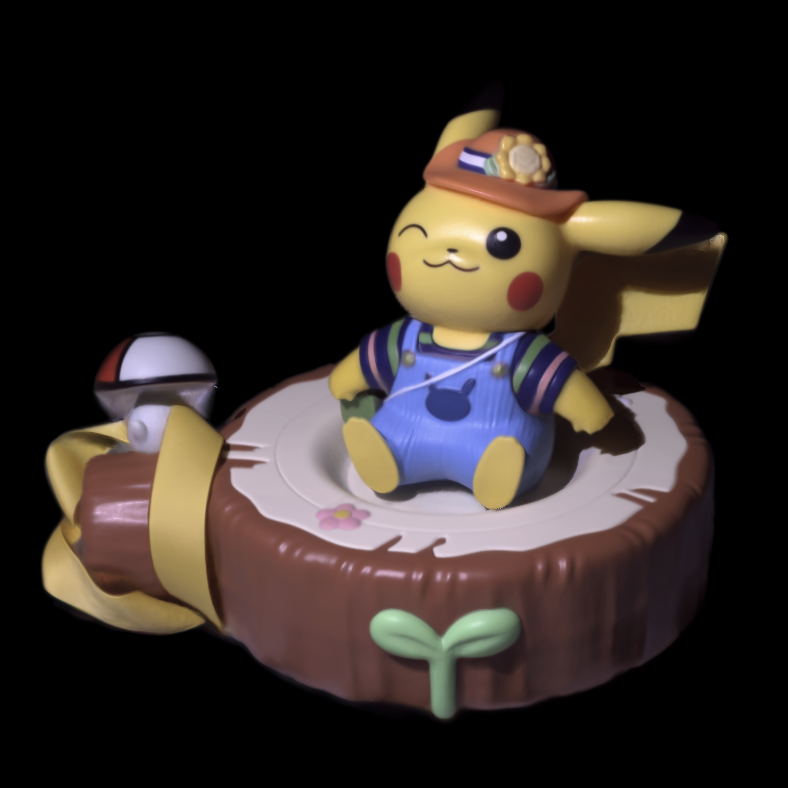}
  &\includegraphics[height=\teaserImgWidth]{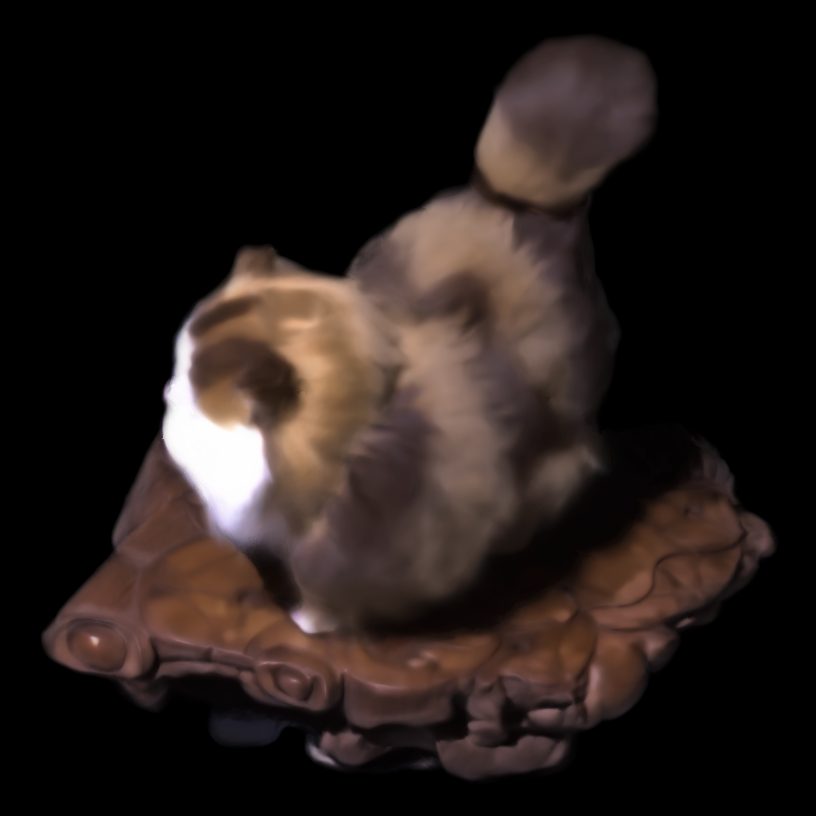}
  &\includegraphics[height=\teaserImgWidth]{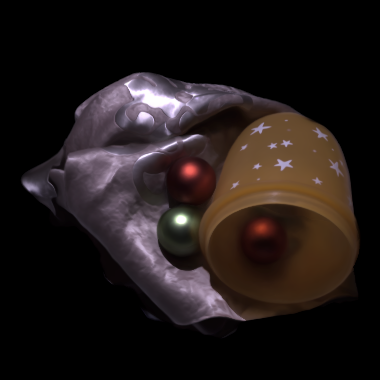}
  &\includegraphics[height=\teaserImgWidth]{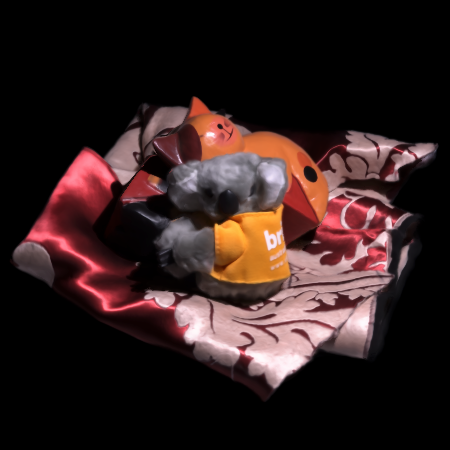}
  &\includegraphics[height=\teaserImgWidth]{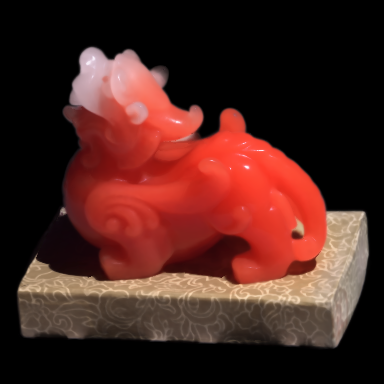}
  &\includegraphics[height=\teaserImgWidth]{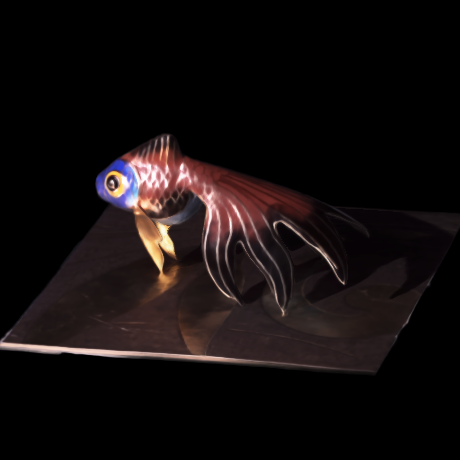}
\end{tabular}
\vspace{-3mm}
\caption{Free viewpoint relighting of neural radiance fields trained on
  $500\!-\!1,\!000$ unstructured photographs per scene captured with a
  handheld setup.}
  \label{fig:teaser} 
\end{teaserfigure}

\maketitle

%!TEX root = ../NeuralRelightHint.tex

\section{Introduction}
The appearance of real-world objects is the result of complex light transport
interactions between the lighting and the object's intricate geometry and
associated material properties. Digitally reproducing the appearance of
real-world objects and scenes has been a long-standing goal in computer
graphics and computer vision.  Inverse rendering methods attempt to undo the
complex light transport to determine a sparse set of model parameters that,
together with the chosen models, replicates the appearance when rendered.
However, teasing apart the different entangled components is ill-posed and
often leads to ambiguities. Furthermore, inaccuracies in one model can
adversely affect the accuracy at which other components can be disentangled,
thus requiring strong regularization and assumptions.

In this paper we present a novel, NeRF-inspired~\cite{Mildenhall:2020:NRS},
neural implicit radiance representation for free viewpoint relighting of
general objects and scenes. Instead of using analytical reflectance models and
inverse rendering of the neural implicit representations, we follow a
data-driven approach and refrain from decomposing the appearance in different
light transport components.  Therefore, unlike the majority of prior work in
relighting neural implicit
representations~\cite{Srinivasan:2021:NNR,Boss:2021:NNR,Kuang:2022:NNR,Boss:2022:SSM,Zheng:2021:NRP},
we relax and enrich the lighting information embedded in handheld captured
photographs of the object by illuminating each view from a random point light
position. This provides us with a broader unstructured sampling of the space
of appearance changes of an object, while retaining the convenience of
handheld acquisition. Furthermore, to improve the reproduction quality of
difficult to learn components, we provide shadow and highlight \emph{hints} to
the neural radiance representation. Critically, we do not impose how these
hints are combined with the estimated radiance (\eg\!, shadow mapping by
multiplying with the light visibility), but instead leave it up to the neural
representation to decide how to incorporate these hints in the final result.

Our hint-driven implicit neural representation is easy to implement, and it
requires an order of magnitude less photographs than prior relighting methods
that have similar capabilities, and an equal number of photographs compared to
state-of-the-art methods that offer less flexibility in the shape and/or
materials that can be modeled.  Compared to fixed lighting implicit
representations such as NeRF~\cite{Mildenhall:2020:NRS}, we only require a
factor of five times more photographs and twice the render cost while gaining
relightability.  We demonstrate the effectiveness and validate the robustness
of our representation on a variety of challenging synthetic and real objects
(\eg\!, \autoref{fig:teaser}) containing a wide range of materials (\eg\!,
subsurface scattering, rough specular materials, etc.) variations in shape
complexity (\eg\!, thin features, ill-defined furry shapes, etc.) and global
light transport effects (\eg\!, interreflections, complex shadowing, etc.).

%!TEX root = ../NeuralRelightHint.tex

\section{Related Work}
\label{sec:related}

We focus the discussion of related work on seminal and recent work in
image-based relighting, inverse rendering, and relighting neural implicit
representations.  For an in-depth overview we refer to recent surveys in
neural rendering~\cite{Tewari:2022:ANR},
(re)lighting~\cite{Einabadi:2021:DNM}, and appearance
modeling~\cite{Dong:2019:DAM}.

\paragraph{Image-based Relighting}
The staggering advances in machine learning in the last decade have also had a
profound effect on image-based relighting~\cite{Debevec:2000:ARF}, enabling
new capabilities and improving
quality~\cite{Ren:2015:IBR,Xu:2018:DIR,Bemana:2020:XFI}.  Deep learning has
subsequently been applied to more specialized relighting tasks for
portraits~\cite{Sun:2019:SIP,Pandey:2021:TRL,Meka:2019:DRF,Bi:2021:DRA,Sun:2020:LSS},
full
bodies~\cite{Meka:2020:DRT,Zhang:2021:NLT,Guo:2019:RVP,Kanamori:2018:RHO,Yeh:2022:LRP},
and outdoor scenes~\cite{Mahmoud:2019:NRW,Philip:2019:MRU,Griffiths:2022:OOS}.
It is unclear how to extend these methods to handle scenes that contain
objects with ill-defined shapes (\eg\!, fur) and translucent and specular
materials.

Our method can also be seen as a free-viewpoint relighting method that
leverages highlight and shadow hints to help model these challenging effects.
Philip~\etal~\shortcite{Philip:2019:MRU} follow a deep shading
approach~\cite{Nalbach:2017:DSC} for relighting, mostly diffuse, outdoor
scenes under a simplified sun+cloud lighting model.  Relit images are created
in a two stage process, where an input and output shadow map computed from a
proxy geometry is refined, and subsequently used, together with additional
render buffers, as input to a relighting network.
Zhang~\etal~\shortcite{Zhang:2021:NLT} introduce a semi-parametric model with
residual learning that leverages a diffuse parametric model (\ie\!, radiance
hint) on a rough geometry, and a learned representation that models
non-diffuse and global light transport embedded in texture space. To
accurately model the non-diffuse effects, Zhang~\etal require a large number
($\sim\!8,\!000$) of structured photographs captured with a light stage.
Deferred Neural Relighting~\cite{Gao:2020:DNL} is closest to our method in
terms of capabilities; it can perform free-viewpoint relighting on objects
with ill-defined shape with full global illumination effects and complex
light-matter interactions (including subsurface scattering and fur). Similar
to Zhang~\etal~\shortcite{Zhang:2021:NLT}, Gao~\etal embed learned features in
the texture space of a rough geometry that are projected to the target view
and multiplied with \emph{radiance cues}. These radiance cues are
visualizations of the rough geometry with different BRDFs (\ie\!, diffuse and
glossy BRDFs with $4$ different roughnesses) under the target lighting with
global illumination.  The resulting images are then used as guidance hints for
a neural renderer trained per scene from a large number ($\sim\!10,\!000$) of
unstructured photographs of the target scene for random point light-viewpoint
combinations to reproduce the reference appearance.
Philip~\etal~\shortcite{Philip:2021:FVI} also use radiance hints (limited to
diffuse and mirror radiance) to guide a neural renderer.  However, unlike
Zhang~\etal and Gao~\etal, they pretrain a neural renderer that does not
require per-scene fine-tuning, and that takes radiance cues for both the input
and output conditions. Philip~\etal require about the same number as input
images as our method, albeit lit by a single fixed natural lighting conditions
and limited to scenes with hard surfaces and BRDF-like materials.  All four
methods rely on multi-view stereo which can fail for complex scenes.  In
contrast our method employs a robust neural implicit representation.
Furthermore, all four methods rely on an image-space neural renderer to
produce the final relit image.  In contrast, our method provides the hints
during volume rendering of the neural implicit representation, and thus it is
independent of view-dependent image contexts.  Our method can relight scenes
with the same complexity as Gao~\etal~\shortcite{Gao:2020:DNL} while only
using a similar number of input photographs as
Philip~\etal~\shortcite{Philip:2021:FVI} without sacrificing robustness.

\paragraph{Model-based Inverse Rendering}
An alternative to data-driven relighting is inverse rendering (a.k.a.
analysis-by-synthesis) where a set of trial model parameters are optimized
based on the difference between the rendered model parameters and reference
photographs.  %Early work in inverse rendering required carefully calibrated
%reference photographs of an object (e.g.,~\cite{Holroyd:2010:COS}). Subsequent
%work has relaxed acquisition requirements to specialized portable
%setups~\cite{Ma:2021:FFS}, colocated camera-light setups~\cite{Nam:2018:PSA},
%polarized lighting/cameras~\cite{Hwang:2022:SEp,Tunwattanapong:2013:ARS}, and
%natural lighting~\cite{Xia:2016:RSS}.
%
Inverse rendering at its core is a complex non-linear optimization
problem. Recent advances in differentiable
rendering~\cite{Nimier-David:2019:M2R,Li:2018:DMC,Loper:2014:OAA,Xing:2022:DRR}
have enabled more robust inverse rendering for more complex scenes and capture
conditions. BID-R++~\cite{Chen:2021:BLP} combines differentiable ray tracing
and rasterization to model spatially varying reflectance parameters and
spherical Gaussian lighting for a known triangle mesh.
Munkberg~\etal~\shortcite{Munkberg:2022:ET3} alternate between optimizing an
implicit shape representation (\ie\!, a signed distance field), and reflectance
and lighting defined on a triangle
mesh. Hasselgren~\etal~\shortcite{Hasselgren:2022:SLM} extend the work of
Munkberg~\etal~\shortcite{Munkberg:2022:ET3} with a differentiable Monte Carlo
renderer to handle area light sources, and embed a denoiser to mitigate the
adverse effects of Monte Carlo noise on the gradient computation to drive the
non-linear optimizer. Similarly, Fujun~\etal~\shortcite{Fujun:2021:USS} also
employ a differentiable Monte Carlo renderer for estimating shape and
spatially-varying reflectance from a small set of colocated view/light
photographs.  All of these methods focus on direct lighting only and can
produce suboptimal results for objects or scenes with strong interreflections.
A notable exception is the method of Cai~\etal~\shortcite{Cai:2022:PBI} that
combines explicit and implicit geometries and demonstrates inverse rendering
under known lighting on a wide range of opaque objects while taking indirect
lighting in account.  All of the above methods eventually express the shape as
a triangle mesh, limiting their applicability to objects with well defined
surfaces.  Furthermore, the accuracy of these methods is inherently limited by
the representational power of the underlying BRDF and lighting models.

%\cite{Knodt:2021:NRT} Neural Ray Tracing  ARXIV
%\cite{Sang:2020:SSN}: joint training of SVBRDF recovery and relighting
%\cite{Bi:2020:D3C}: learning based no indirect
%!TEX root = ../NeuralRelightHint.tex
\begin{figure*}[th]
  \includegraphics[width=\textwidth]{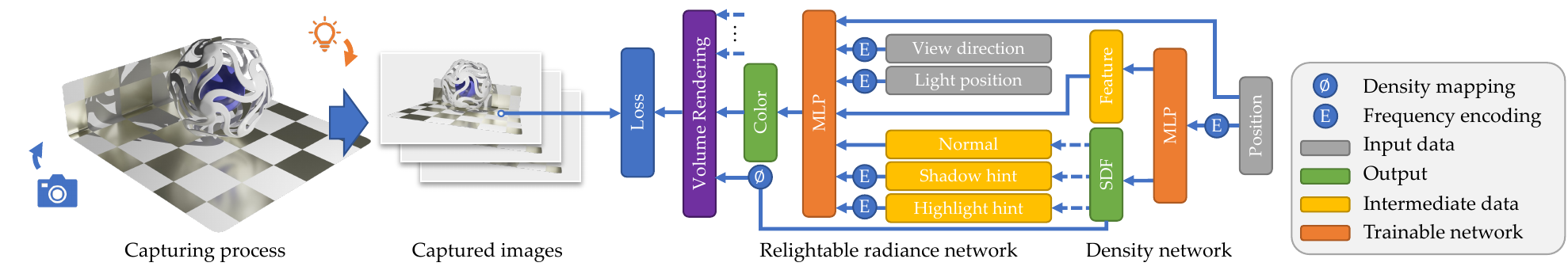}
  \caption{Overview: our neural implicit radiance representation is trained on
    unstructured photographs of the scene captured from different viewpoints
    and lit from different point light positions.  The neural implicit
    radiance representation consists of two multi layer perceptron (MLP)
    networks for modeling the density field and for modeling the light
    transport.  The MLP for modeling the density takes as input the position,
    and outputs the signed distance function of the shape and a feature vector
    that together with the current position, the normal extracted from the
    SDF, the view direction, the light source position, and the light
    transport hints, are passed into the radiance MLP that then computes
    the view and lighting dependent radiance.}
  \label{fig:overview}
\end{figure*}

\paragraph{Neural Implicit Representations}
A major challenge in inverse rendering with triangle meshes is to efficiently
deal with changes in topology during optimization.  An alternative to triangle
mesh representations is to use a volumetric representation where each voxel
contains an opacity/density estimate and a description of the reflectance
properties.  While agnostic to topology changes, voxel grids are memory
intensive and, even with grid warping~\cite{Bi:2020:DRV}, fine-scale
geometrical details are difficult to model.

To avoid the inherent memory overhead of voxel grids, NeRF
\cite{Mildenhall:2020:NRS} models the continuous volumetric density and
spatially varying color with two multi layer perceptrons (MLPs) parameterized
by position (and also view direction for color). The MLPs in NeRF are trained
per scene such that the accumulated density and color ray marched along a view
ray matches the observed radiance in reference photographs.  NeRF has been
shown to be exceptionally effective in modeling the outgoing radiance field of
a wide range of object types, including those with ill-defined shapes and
complex materials. % (e.g., translucent, transparent, hair, etc.).
One of the main limitations of NeRF is that the illumination present at
capture-time is baked into the model.  Several methods have been introduced to
support post-capture relighting under a restricted lighting
model~\cite{martinbrualla:2020:NWN,Li:2022:NLF}, or by altering the color MLP
to produce the parameters to drive an analytical model of the appearance of
objects~\cite{Zhang:2021:NFS,Boss:2021:NNR,Boss:2021:NPN,Boss:2022:SSM,Srinivasan:2021:NNR,Yao:2022:NNI,Kuang:2022:NNR},
participating media~\cite{Zheng:2021:NRP}, or even whole outdoor
scenes~\cite{Rudnev:2022:NOS}.

Due to the high computational cost of ray marching secondary rays, na\"ively
computing shadows and indirect lighting is impractical.
Zhang~\etal\shortcite{Zhang:2021:NFS}, Li~\etal~\shortcite{Li:2022:NLF}, and
Yang~\etal~\shortcite{Yang:2022:PNI} avoid tracing shadow rays by learning an
additional MLP to model the ratio of light occlusion. However, all three
methods ignore indirect lighting.  Zheng~\etal~\shortcite{Zheng:2021:NRP}
model the indirect lighting inside a participating media using an MLP that
returns the coefficients of a 5-band expansion. NeILF~\cite{Yao:2022:NNI}
embeds the indirect lighting and shadows in a (learned) 5D incident light
field for a scene with \emph{known} geometry. NeRV~\cite{Srinivasan:2021:NNR}
modifies the color MLP to output BRDF parameters and a visibility field that
models the distance to the nearest 'hard surface' and lighting visibility.
The visibility field allows them to bypass the expensive ray marching step for
shadow computation and \emph{one-bounce} indirect illumination.  A
disadvantage of these solutions is that they do not guarantee that the
estimated density field and the occlusions are coupled.  In contrast, our
method directly ties occlusions to the estimated implicit geometry reproducing
more faithful shadows. Furthermore, these methods rely on BRDFs to model the
surface reflectance, precluding scenes with complex light-matter interactions.

NeLF~\cite{Sun:2021:NNL} aims to relight human faces, and thus accurately
reproducing subsurface scattering is critical. Therefore,
Sun~\etal characterize the radiance and global light transport by an MLP.  We
also leverage an MLP to model local and global light transport.  A key
difference is that our method parameterizes this MLP in terms of view and
light directions, whereas NeLF directly outputs a full light transport vector
and compute a relit color via an inner-product with the lighting. While better
suited for relighting with natural lighting, NeLF is designed for relighting
human faces which only exhibit limited variations in shape and reflectance.

Similar in spirit to our method, Lyu~\etal~\shortcite{Lyu:2022:NRT} model
light transport using an MLP, named a Neural Radiance Transfer Field (NRTF).
However, unlike us, Lyu~\etal~ train the MLP on synthetic training data
generated from a rough BRDF approximation obtained through physically based
inverse rendering on a triangle mesh extracted from a neural signed distance
field~\cite{Wang:2021:NLN} computed from unstructured observations of the
scene under static natural lighting. To correct the errors due the rough BRDF
approximation, a final refinement step of the MLP is performed using the
captured photographs.  Similar to Lyu~\etal~ we also use an MLP to model light
transport, including indirect lighting.  However, unlike Lyu~\etal we do not
rely solely on an MLP to model high frequency light transport effects such as
light occlusions and specular highlights. Instead we provide shadow and
highlight hints to the radiance network and let the training
process discover how to best leverage these hints.  Furthermore, we rely on a
neural representation for shape jointly optimized with the radiance,
allowing us to capture scenes with ill-defined geometry. In contrast,
Lyu~\etal optimize shape (converted to a triangle mesh) and radiance
separately, making their method sensitive to shape errors and restricted to
objects with a well-defined shape.

An alternative to using an implicit neural density field, is to model the
shape via a signed distance field (SDF).  Similar to the majority of
NeRF-based methods, PhySG~\cite{Zhang:2021:PIR} and IRON~\cite{Zhang:2022:IRO}
also rely on an MLP to represent volumetric BRDF parameters. However, due to
the high computational cost, these methods do not take shadowing or indirect
lighting in account.  Zhang~\etal~\shortcite{Zhang:2022:MII} model indirect
lighting separately, and train an additional incident light field MLP using
the incident lighting computed at each point via ray casting the SDF geometry.
While our method also builds on a neural implicit
representation~\cite{Wang:2021:NLN}, our method does not rely on an underlying
parametric BRDF model, but instead models the full light transport via an MLP.
Furthermore, we do not rely on an MLP decoupled from the estimated geometry to
estimate shadowing, but instead accumulate light occlusion along a single
shadow ray per view ray, ensuring consistency between the shadows and the
estimated geometry.

%!TEX root = ../NeuralRelightHint.tex

\section{Method}
\label{sec:method}
Our goal is to extend neural implicit representations such as NeRF
\cite{Mildenhall:2020:NRS} to model variations in lighting.  NeRF has
proven to be exceptionally efficient for viewpoint interpolation. %, the
%underlying implicit volumetric representation is intrinsically inefficient for
%computing secondary rays.
In contrast to ray tracing with solid surfaces, NeRF relies on ray marching
through the volume, requiring at least an order of magnitude more
computations. Not only does this ray marching cost affect rendering, it also
leads to a prohibitively large training cost when secondary rays (\eg\!,
shadows and indirect lighting) are considered.  Instead of building our method
on NeRF, we opt for using NeuS~\cite{Wang:2021:NLN}, a neural implicit signed
distance field representation, as the basis for our method. Although NeuS does
not speed up ray marching, it provides an unbiased depth estimate which we
will leverage in~\autoref{sec:hints} for reducing the number of shadow rays.

Following prior work, our neural implicit radiance representation relies on
two multi layer perceptrons (MLPs) for modeling the density field (following
NeuS) and for modeling the (direct and indirect) radiance based on the current
position, the normal derived from the density field, the view direction, the
point light position, and the features provided by the density network. In
addition, we also provide light transport \emph{hints} to the relightable
radiance MLP to improve the reproduction quality of difficult to model effects
such as shadows and highlights. \autoref{fig:overview} summarizes our
architecture.

To train our neural implicit relightable radiance representation, we require
observations of the target scene seen from different viewpoints and lit from
different point light positions. It is essential that these observations
include occlusions and interreflections.  Colocated lighting (\eg\!, as
in~\cite{Nam:2018:PSA,Fujun:2021:USS}) does not exhibit visible shadows and is
therefore not suited.  Instead we follow the acquisition process of Deferred
Neural Lighting~\cite{Gao:2020:DNL} and capture the scene from different
viewpoints with a handheld camera while lighting the scene with a flash light
of a second camera from a different direction.

We opt for parameterizing the radiance function with respect to a point light
as the basis for relighting as this better reflects the physical capture
process. A common approximation in prior religting work that relies on active
illumination (\eg\!, Light Stage) is to ignore the divergence of incident
lighting due to the finite light source distance, and parameterize the
reflectance field in terms lighting directions only. Similarly, we can also
\emph{approximate} distant lighting with point lighting defined by projecting
the light direction onto a large sphere with a radius equal to the capture
distance.

%The following subsections detail our neural implicit relightable radiance
%representation, the light transport hints, training and loss, and data
%acquisition; \autoref{fig:overview} summarizes our method and architecture.

\subsection{Representation}

\paragraph{Density Network}
Our neural implicit geometry representation follows NeuS~\cite{Wang:2021:NLN}
which uses an MLP to encode a Signed Distance Function (SDF) $\sdf$ from which
the density function is derived using a probability density function
$\phi_s(\sdf)$.  This probability density function is designed to ensure that
for opaque objects the zero-level set of the SDF corresponds to the
surface. The width of the probability distribution models the uncertainty of
the surface location.  We follow exactly the same architecture for the density
MLP as in NeuS: $8$ hidden layers with $256$ nodes using a Softplus activation
and a skip connection between the input and the $4$th layer.  The input (\ie\!,
current position along a ray) is augmented using a frequency encoding with
$6$ bands.
%
%\begin{eqnarray}
%  \! \gamma_L(\pos) = (\pos, \sin(2^0\pi \pos), \cos(2^0\pi \pos), ...,
%  \sin(2^{L-1}\pi \pos), \cos(2^{L-1} \pi \pos)); \!\!\!\!
%\label{eq:encoding}
%\end{eqnarray}
%
%note that
In addition, we also concatenate the original input signal to the encoding.
The resulting output from the density network is the SDF at $\pos$ as well as
a latent vector that encodes position dependent features.

\paragraph{Relightable Radiance  Network}
Analogous to the color MLP in NeRF and NeuS that at each volumetric position
evaluates the view-dependent color, we introduce a \emph{relightable radiance}
MLP that at each volumetric position evaluates the view and lighting dependent
(direct and indirect) light transport.  We follow a similar architecture as
NeRF/NeuS' color MLP and extend it by taking the position dependent feature
vector produced by the density MLP, the normal derived from the SDF, the
current position, the view direction, and the point light position as
input. Given this input, the radiance MLP outputs the resulting radiance which
includes all light transport effects such as occlusions and
interreflections. We assume a white light source color; colored lighting can
be achieved by scaling the radiance with the light source color (\ie\!,
linearity of light transport).

Given the output from the density network $\f$ as well as the output from the
radiance network $\s$, the color $\C$ along a view ray starting at the
camera position $\camera$ in a direction $\viewDir$ is given by:
\begin{eqnarray}
  \label{eq:render}
  \C(\camera, \viewDir) = \int_0^\infty \w(t) \s(\pos, \normal, \viewDir, \lightPos, \feat, \cues) \intd t,
\end{eqnarray}
where the sample position along the view ray is $\pos = \camera + t \viewDir$
at depth $t$, $\normal$ is the normal computed as the normalized SDF gradient:
\begin{eqnarray}
\label{eq:normal}
	\normal = \nabla \sdf  / || \nabla \sdf ||, 
\end{eqnarray}
$\viewDir$ is the view direction, $\lightPos$ is the point light position,
$\feat$ the corresponding feature vector from the density MLP, and $\cues$ is
a set of additional hints provided to the radiance network (described
in~\autoref{sec:hints}). Analogous to NeuS, the view direction, light
position, and hints are all frequency encoded with $4$ bands. %; we omit this
%encoding from~\autoref{eq:render} for brevity.
%
Finally, $\w(t)$ is the unbiased density weight~\cite{Wang:2021:NLN} computed
by:
\begin{eqnarray}
  \w(t) & = & T(t) \rho(t), \\
  T(t)  & = & \exp \left(-\int_0^t \rho(u) \intd u \right), \\
  \rho(t) & = & \max \left( \frac{\frac{\intd \Phi_s}{\intd t}(\f(t))}{\Phi_s(\f(t))}, 0 \right), %\\
%  \Phi_s(x) & = & \int_0^t \phi_s(t) \intd t,
\end{eqnarray}
with $T$ the transmittance over opacity $\rho$, $\Phi_s$ the CDF of the PDF
$\phi_s$ used to compute the density from the SDF $\f$.  To speed up the
computation of the color, the integral in~\autoref{eq:render} is computed by
importance sampling the density field along the view ray.

In the spirit of image-based relighting, we opt to have the relightable
radiance MLP network include global light transport effects such as
interreflections and occlusions.  While MLPs are in theory universal
approximators, some light transport components are easier to learn (\eg\!,
diffuse reflections) than others. Especially high frequency light transport
components such as shadows and specular highlights pose a problem.  At the
same time, shadows and specular highlights are highly correlated with the
geometry of the scene and thus the density field.  To leverage this embedded
knowledge, we provide the relightable radiance MLP with additional
\emph{shadow} and \emph{highlight hints}.  %While we leverage prior knowledge
%from rendering to compute these hints, we leave it to the network to learn how
%to incorporate these hints in the computation of the radiance.

\subsection{Light Transport Hints}
\label{sec:hints}

\paragraph{Shadow Hints}
While the relightable radiance network is able to roughly model the effects of
light source occlusion, the resulting shadows typically lack sharpness and
detail. Yet, light source occlusion can be relatively easily evaluated by
collecting the density along a shadow ray towards the light source. While this
process is relatively cheap for a single shadow ray, performing a secondary
ray march for each primary ray's sampled position increases the computation
cost by an order of magnitude, quickly becoming too expensive for practical
training.  However, we observe that for most primary rays, the ray samples are
closely packed together around the zero level-set in the SDF due to the
importance sampling of the density along the view ray.  Hence, we propose to
approximate light source visibility by shooting a single shadow ray at the
zero level-set, and use the same light source visibility for each sample along
the view ray. To determine the depth of the zero level-set, we compute the
density weighted depth along the view ray:
\begin{eqnarray}
  \label{eq:volumedepth}
  \D(\camera, \viewDir) = \int_0^{\infty} \w(\pos) t \intd t. 
\end{eqnarray}
\ignorethis{
%% The following is really an implementation detail that comes out of the blue
%% without any context.  I think it is better to leave this out.
In practice we found that the depth by ray marching along the view ray
using~\autoref{eq:volumedepth} has a RMSE smaller than $7e-4$ compared to the
standard sphere tracing method for computing depth, while at the same time
being $2770$ times faster.
}

While for an opaque surface a single shadow ray is sufficient, for non-opaque
or ill-defined surfaces a single shadow ray offers a poor estimate of the
light occlusion.  Furthermore, using the shadow information as a hard mask,
ignores the effects of indirect lighting. We therefore provide the shadow
information as a additional input to the radiance network, allowing the
network learn whether to include or ignore the shadowing information as well
as blend any indirect lighting in the shadow regions.

\paragraph{Highlight Hints}
Similar to shadows, specular highlights are spar\-sely distributed high
frequency light transport effects.  Inspired by Gao~\etal~\shortcite{Gao:2020:DNL},
we provide specular highlight hints to the radiance network by evaluating
$4$ microfacet BRDFs with a GGX distribution~\cite{Walter:2007:MMR} with
roughness parameters $\{0.02,$ $0.05,$ $0.13,$ $0.34 \}$. Unlike Gao~\etal,
we compute the highlight hints using local shading which only depends on the
surface normal computed from the SDF (\autoref{eq:normal}), and pass it to the
radiance MLP as an additional input.  Similar to shadow hints, we compute one
highlight hint per view ray and reused it for all samples along the view ray.

%!TEX root = ../../../NeuralRelightHint.tex

\newcommand{\synImgWidth}{0.16\textwidth}

\begin{figure*}
\renewcommand{\arraystretch}{0.3}
\addtolength{\tabcolsep}{-5.5pt}
\begin{tabular}{ cccccc }
        \begin{overpic}[width=\synImgWidth,percent]{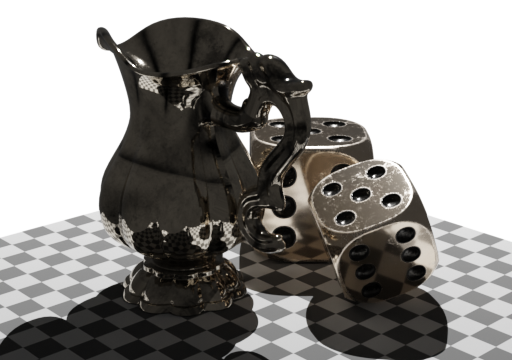}\end{overpic}
        &\begin{overpic}[width=\synImgWidth,percent]{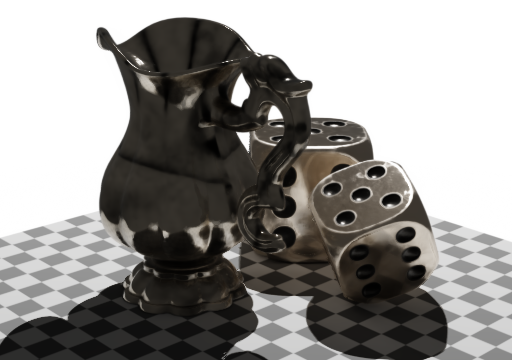}\end{overpic}
        &\begin{overpic}[width=\synImgWidth,percent]{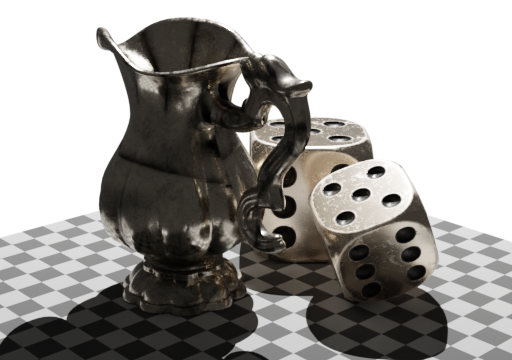}\end{overpic}
        &\begin{overpic}[width=\synImgWidth,percent]{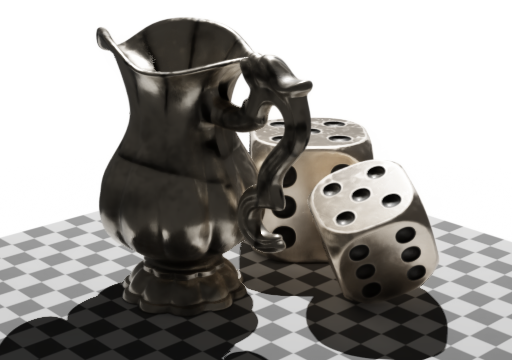}\end{overpic}
        &\begin{overpic}[width=\synImgWidth,percent]{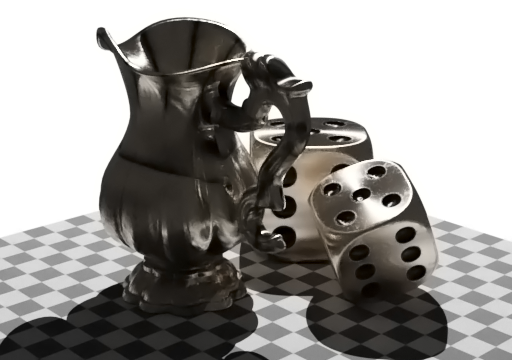}\end{overpic}
        &\begin{overpic}[width=\synImgWidth,percent]{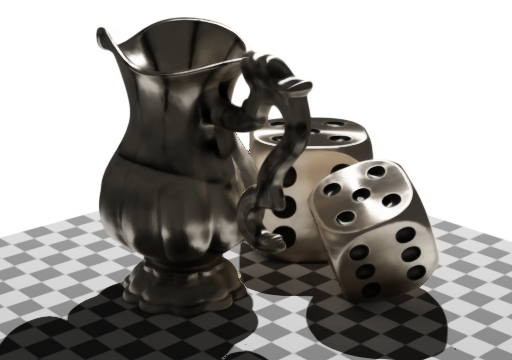}\end{overpic}
        \\
        \multicolumn{2}{c}{\small {\SceneName{Metallic}: 27.79 | 0.9613 | 0.0487}}
        &\multicolumn{2}{c}{\small {\SceneName{Glossy-Metal}: 30.08 | 0.9722 | 0.0376}}
        &\multicolumn{2}{c}{\small {\SceneName{Anisotropic-Metal}: 29.07 | 0.9676 | 0.0395}} \\
        \begin{overpic}[width=\synImgWidth,percent]{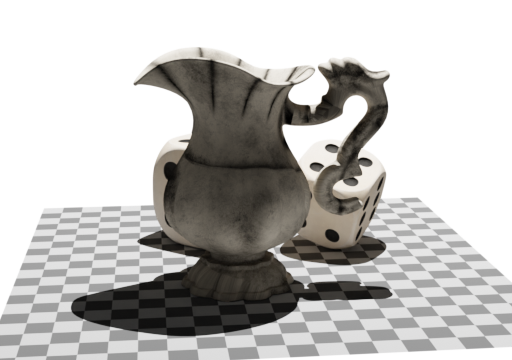}\end{overpic}
        &\begin{overpic}[width=\synImgWidth,percent]{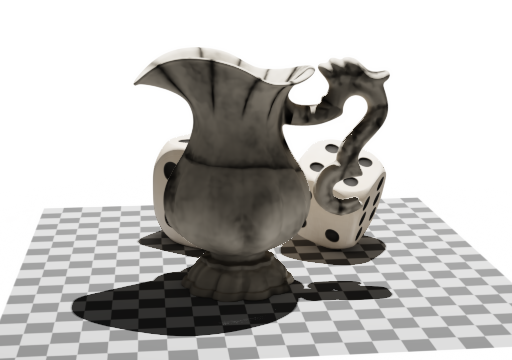}\end{overpic}
        &\begin{overpic}[width=\synImgWidth,percent]{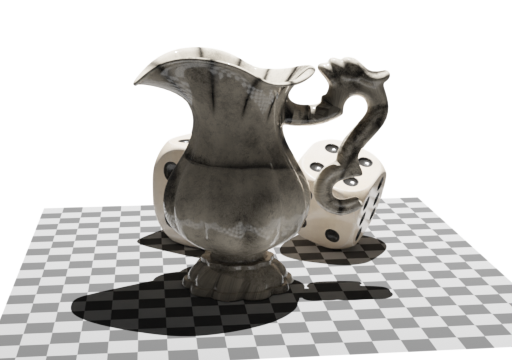}\end{overpic}
        &\begin{overpic}[width=\synImgWidth,percent]{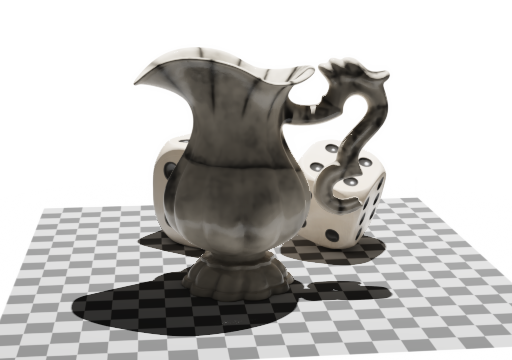}\end{overpic}
        &\begin{overpic}[width=\synImgWidth,percent]{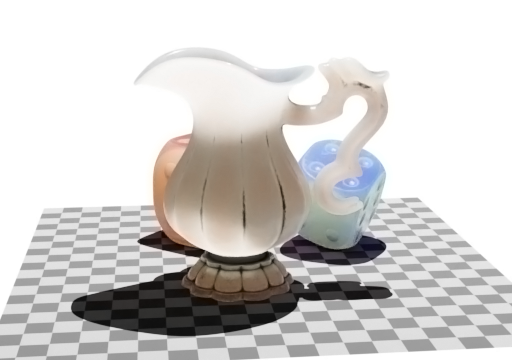}\end{overpic}
        &\begin{overpic}[width=\synImgWidth,percent]{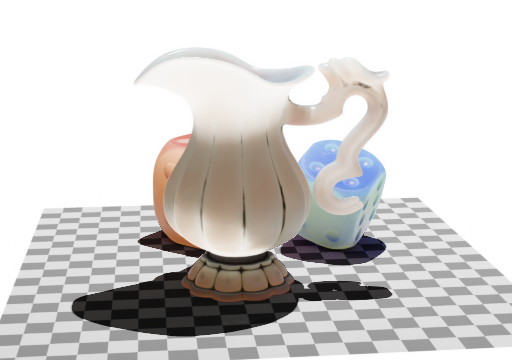}\end{overpic}
        \\
        \multicolumn{2}{c}{\small {\SceneName{Diffuse} 37.10 | 0.9942 | 0.0136}} 
        &\multicolumn{2}{c}{\small {\SceneName{Plastic}: 34.94 | 0.9885 | 0.0210}}
        &\multicolumn{2}{c}{\small {\SceneName{Translucent}: 36.22 | 0.9911 | 0.0172  }}
        \\
        \begin{overpic}[width=\synImgWidth,percent]{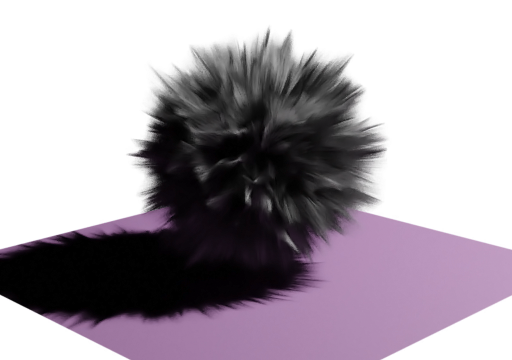}\end{overpic}
        &\begin{overpic}[width=\synImgWidth,percent]{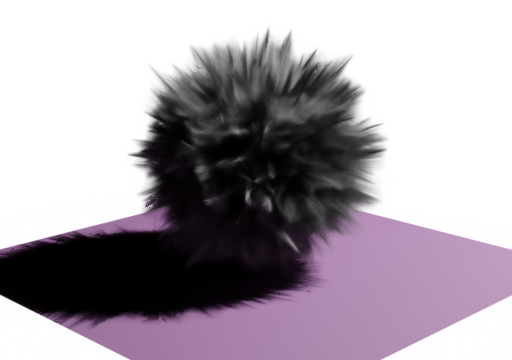}\end{overpic}
        &\begin{overpic}[width=\synImgWidth,percent]{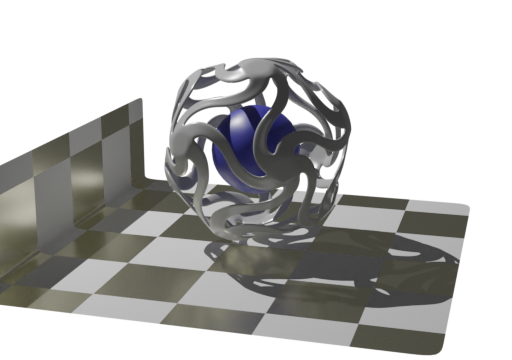}\end{overpic}
        &\begin{overpic}[width=\synImgWidth,percent]{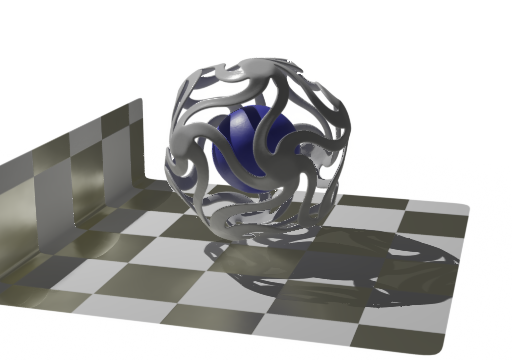}\end{overpic}
        &\begin{overpic}[width=\synImgWidth,percent]{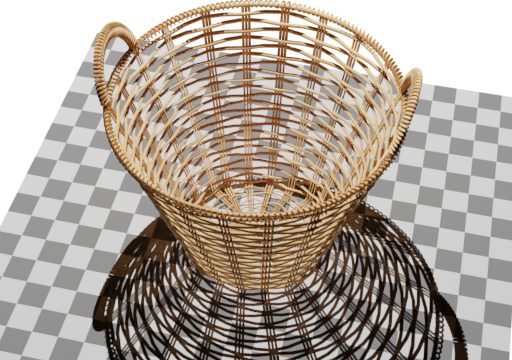}\end{overpic}
        &\begin{overpic}[width=\synImgWidth,percent]{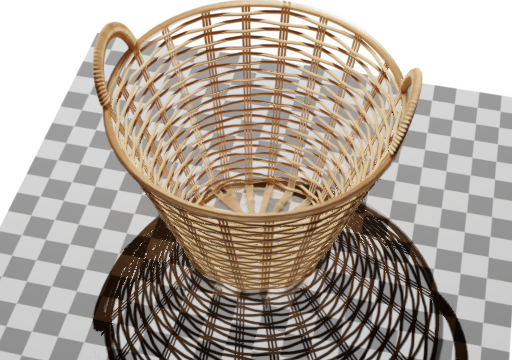}\end{overpic}
        \\
        \multicolumn{2}{c}{\small {\SceneName{Fur Ball}: 32.18 | 0.9619 | 0.0613}} 
        &\multicolumn{2}{c}{\small {\SceneName{Layered Woven Ball} | 33.52 | 0.9853 | 0.0209}} 
        &\multicolumn{2}{c}{\small {\SceneName{Basket}: 26.84 | 0.9586 | 0.0411}}
        \\
        \begin{overpic}[width=\synImgWidth,percent]{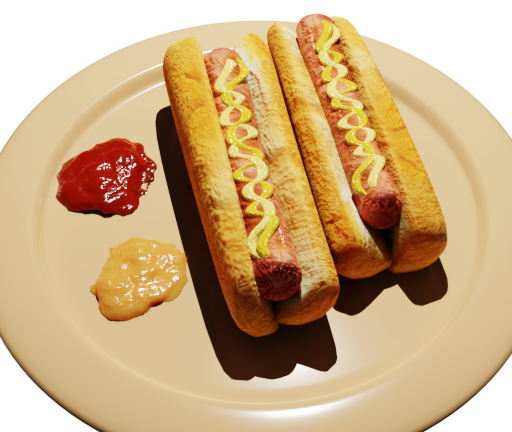}\end{overpic}
        &\begin{overpic}[width=\synImgWidth,percent]{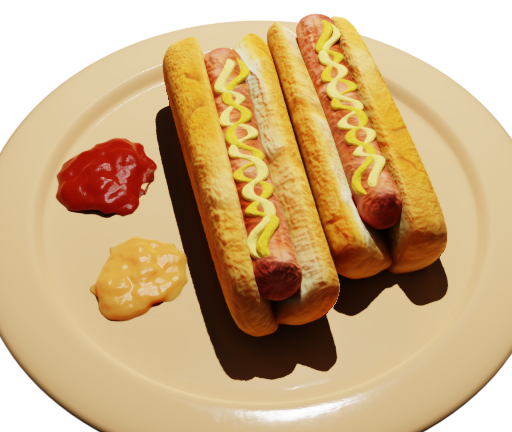}\end{overpic}
        &\begin{overpic}[width=\synImgWidth,percent]{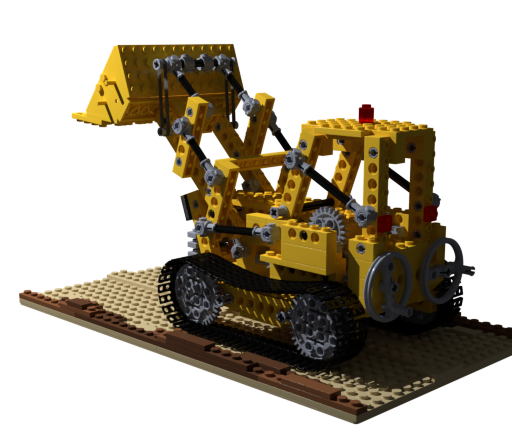}\end{overpic}
        &\begin{overpic}[width=\synImgWidth,percent]{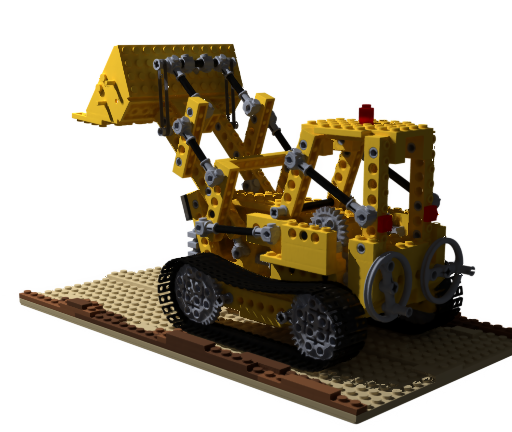}\end{overpic}
        &\begin{overpic}[width=\synImgWidth,percent]{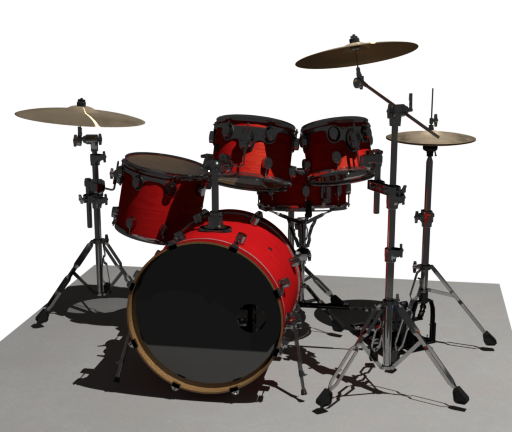}\end{overpic}
        &\begin{overpic}[width=\synImgWidth,percent]{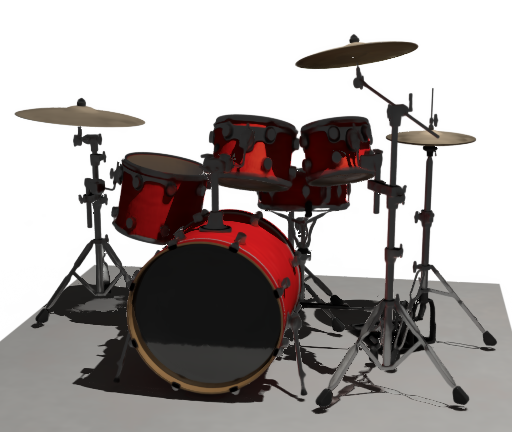}\end{overpic}
        \\
        \multicolumn{2}{c}{\small {\SceneName{Hotdog}: 34.18 | 0.9851 | 0.0246}} 
        &\multicolumn{2}{c}{\small {\SceneName{Lego}: 29.93 | 0.9719 | 0.0301}}
        &\multicolumn{2}{c}{\small {\SceneName{Drums}: 27.92 | 0.9556 | 0.0623}} 
\end{tabular}
\caption{Qualitative comparison between synthetic scenes relit (right) for a novel viewpoint and lighting direction (not part of the training data) and a rendered reference image (left). For each example we list average PSNR, SSIM, and LPIPS computed over a uniform sampling of view and light positions.}
  \label{fig:baseline}
\end{figure*}

\subsection{Loss \& Training}
\label{sec:loss}
We jointly train the density and radiance network using an image
reconstruction loss $\Loss_c$ and an SDF regularization loss $\Loss_e$.  The
image reconstruction loss is defined as the $L_1$ distance between the
observation $\bar{\C}(\camera, \viewDir)$ and the corresponding estimated
color $\C(\camera, \viewDir)$ computed using~\autoref{eq:render}:
$\Loss_{c} = || \bar{C} - C ||_{1}$, for a random sampling of pixels (and thus
view rays) in the captured training images
(\autoref{sec:capture}). Furthermore, we follow NeuS, and regularize the
density MLP with the Eikonal loss~\cite{gropp2020implicit} to ensure a valid
SDF: $\Loss_{e} = ( ||\nabla \sdf||_2 - 1)^2$. For computational efficiency,
we do not back-propagate gradients from the shadow and highlight hints.

\subsection{Data Acquisition}
\label{sec:capture}
Training the implicit representation requires observations of the scene viewed
from random viewpoints and lit from a different random light position such
that shadows and interreflections are included.  We follow the procedure from
Gao~\etal~\shortcite{Gao:2020:DNL}: a handheld camera is used to capture
photographs of the scene from random viewpoints while a second camera captures
the scene with its colocated flash light enabled.  The images from the second
camera are only used to calibrate the light source position. To aid camera
calibration, the scene is placed on a checkerboard pattern.

All examples in this paper are captured with a Sony A7II as the primary
camera, and an iPhone 13 Pro as the secondary camera.  The acquisition process
takes approximately $10$ minutes; the main bottleneck in acquisition is moving
the cameras around the scene.  In practice we capture a video sequence from
each camera and randomly select $500\!-\!1,\!000$ frames as our training data.
The video is captured using S-log encoding to minimize
overexposure. % and color
%corrected after capture using a simple LUT-profile of the camera.  \PP{this
%seems like an unnecessary detail}

For the synthetic scenes, we simulate the acquisition process by randomly
sampling view and light positions on the upper hemisphere around the scene
with a random distance between $2$ to $2.5$ times the size of the scene.  The
synthetic scenes are rendered with global light transport using Blender
Cycles.

\subsection{Viewpoint Optimization}
\label{sec:cam_opt}

Imperfections in camera calibration can cause inaccurate reconstructions
of thin geometrical features as well as lead to blurred results.  To mitigate
the impact of camera calibration errors, we jointly optimize the viewpoints
and the neural representation.

Given an initial view orientation $R_0$ and view position $t_0$, we formulate
the refined camera orientation $R$ and position $t$ as:
\begin{eqnarray}
  R & = & \Delta R \cdot R_0, \\
  t & = & \Delta t + \Delta R \cdot t_0, %\\
\end{eqnarray}
where $\Delta R \in \text{SO(3)}$ and $\Delta t \in \mathbb{R}^3$ are
learnable correction transformations. During training, we back-propagate, the
reconstruction loss, in addition to the relightable radiance network, to the
correction transformations.  We assume that the error on the initial camera
calibration is small, and thus we limit the viewpoint changes by using a
$0.06\times$ smaller learning rate for the correction transformations.

%!TEX root = ../NeuralRelightHint.tex

\section{Results}
\label{sec:results}

%!TEX root = ../../../NeuralRelightHint.tex
\newcommand{\realImgWidth}{0.17\textwidth}
\begin{figure*}
  \renewcommand{\arraystretch}{0.3}
\addtolength{\tabcolsep}{-5.5pt}
\begin{tabular}{ cccccc }
        \begin{overpic}[height=\realImgWidth,percent]{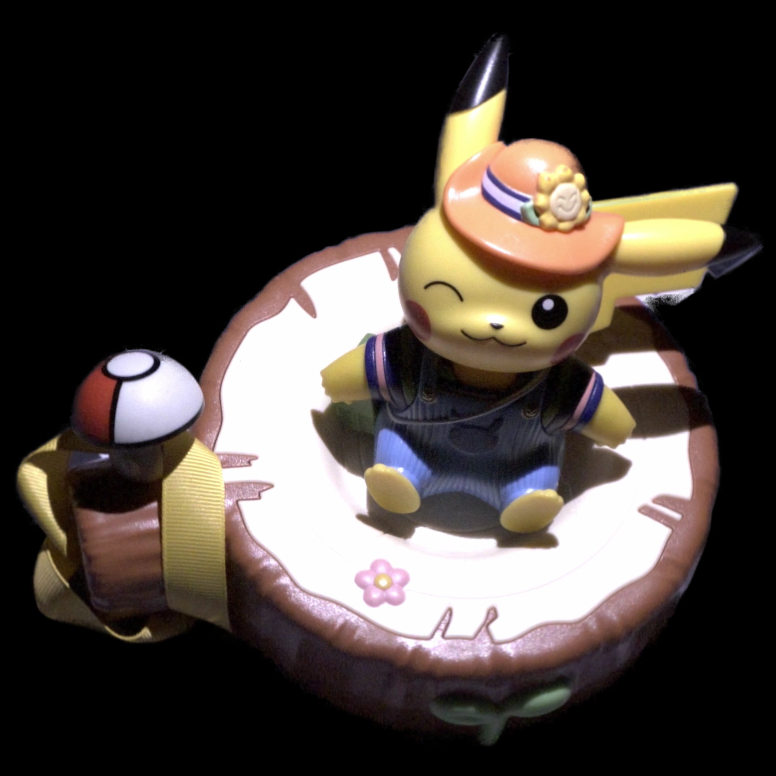}
        \put(3, 90){ \color{white} \small \SceneName{Pikachu statue}}
        \end{overpic}
        &\begin{overpic}[height=\realImgWidth,percent]{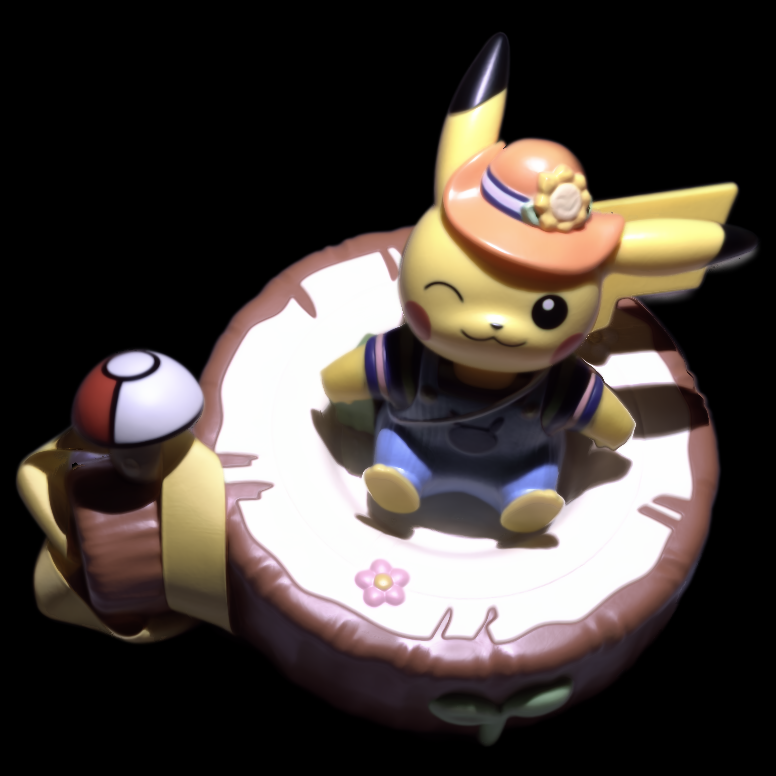}
        \put(3, 90){ \color{white} \footnotesize PSNR: 35.08 }
        \put(3, 83){ \color{white} \footnotesize SSIM: 0.9877}
        \put(3, 76){ \color{white} \footnotesize LPIPS: 0.0359}\end{overpic}         
        &\begin{overpic}[height=\realImgWidth,percent]{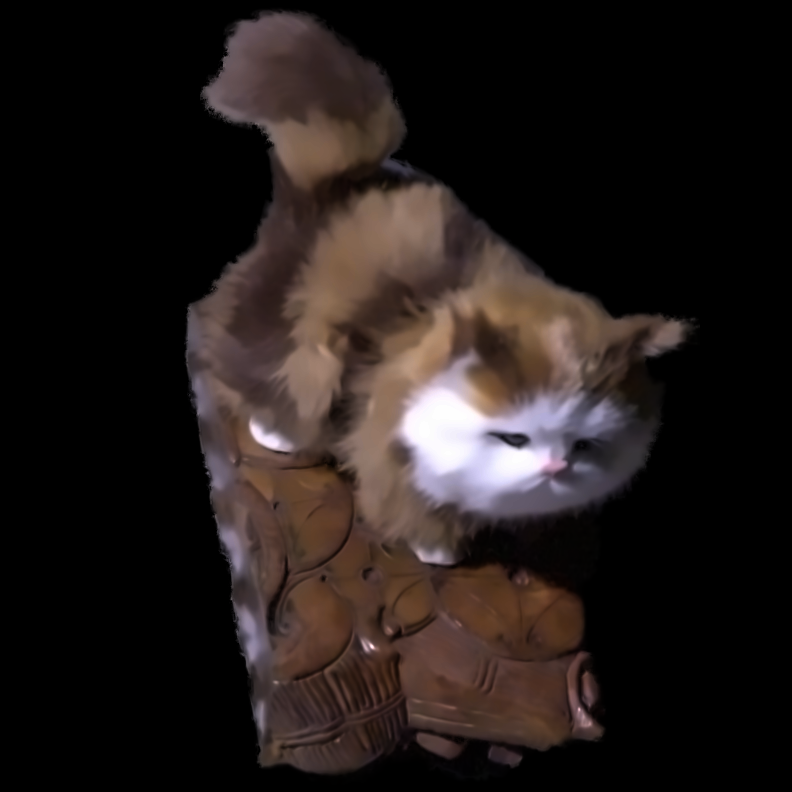}
        \put(3, 90){ \color{white} \small \SceneName{Cat on decor}}\end{overpic}
        &\begin{overpic}[height=\realImgWidth,percent]{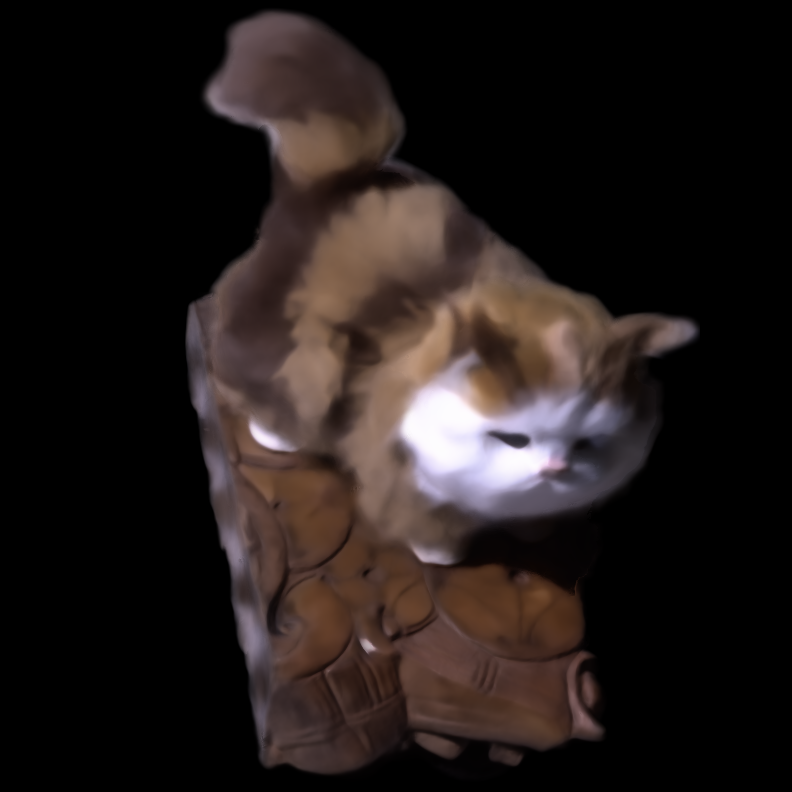}
        \put(-20, 90){ \color{white} \footnotesize PSNR: 36.39 }
        \put(-20, 83){ \color{white} \footnotesize SSIM: 0.9850}
        \put(-20, 76){ \color{white} \footnotesize LPIPS: 0.0604}\end{overpic}
        &\begin{overpic}[height=\realImgWidth,percent]{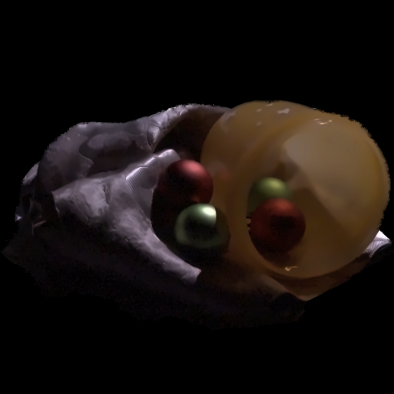}
        \put(3, 90){ \color{white} \small \SceneName{Cup and fabric}}\end{overpic}
        &\begin{overpic}[height=\realImgWidth,percent]{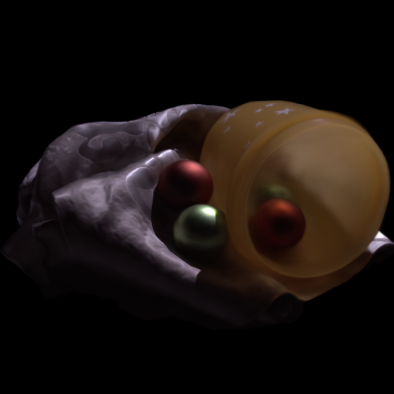}
        \put(-5, 90){ \color{white} \footnotesize PSNR: 38.17 }
        \put(-5, 83){ \color{white} \footnotesize SSIM: 0.9900}
        \put(-5, 76){ \color{white} \footnotesize LPIPS: 0.0355}\end{overpic}
\end{tabular}
\caption{Qualitative comparison between captured scenes relit (right) for a novel viewpoint and lighting direction (not part of the training data) and a reference photograph (left). For each example we list average PSNR, SSIM, and LPIPS computed over randomly sampled view and light positions. }
  \label{fig:real}
\end{figure*}
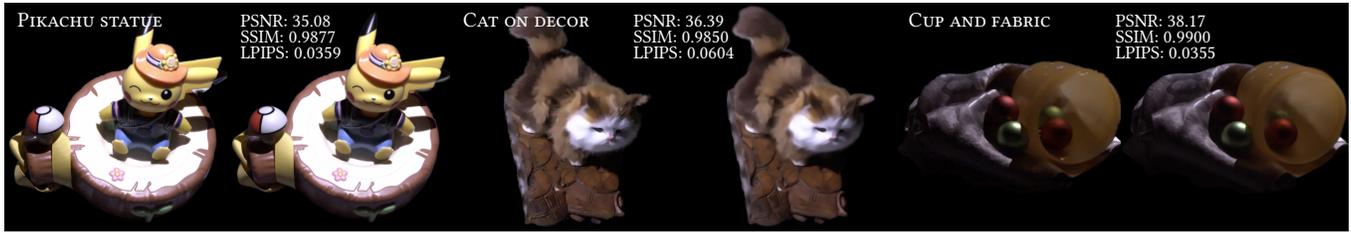

%!TEX root = ../../NeuralRelightHint.tex

\newcommand{\ironFigWidth}{0.23\textwidth}

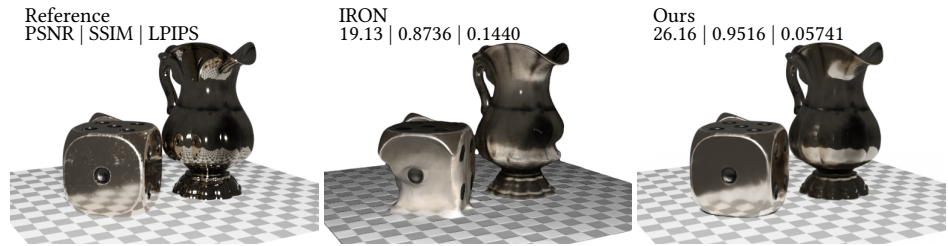
\begin{figure*}
    \begin{minipage}{\ironFigWidth} %
      \centering
        \begin{overpic}[width=\textwidth,percent]{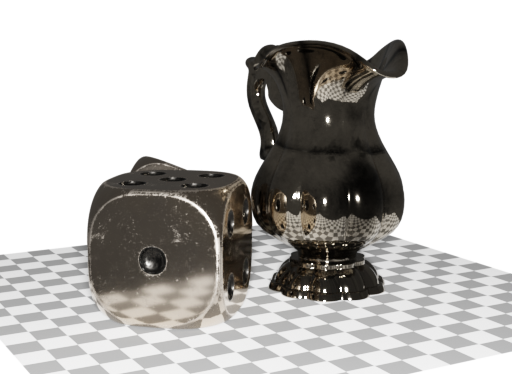}
          \put(3, 72){ \color{black} \small {Reference}}
          \put(3, 66){ \color{black} \small {PSNR | SSIM | LPIPS}}
        \end{overpic}
    \end{minipage} 
    \begin{minipage}{\ironFigWidth} %
        \begin{overpic}[width=\textwidth,percent]{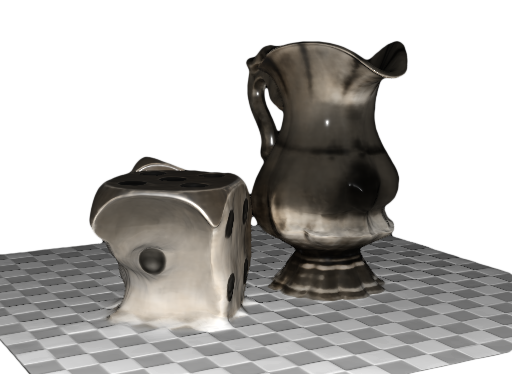}
          \put(3, 72){ \color{black} \small {IRON}}
          \put(3, 66){ \color{black} \small {19.13 | 0.8736 | 0.1440}}
        \end{overpic}
    \end{minipage}
    \begin{minipage}{\ironFigWidth} %
        \begin{overpic}[width=\textwidth,percent]{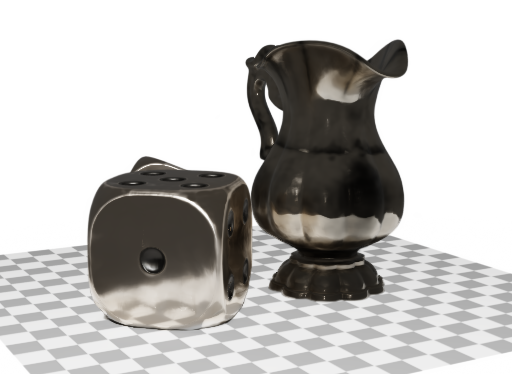}
          \put(3, 72){ \color{black}\small  {Ours}}
          \put(3, 66){ \color{black} \small {26.16 | 0.9516 | 0.05741}}
        \end{overpic}
    \end{minipage} 
    \caption{Comparison to inverse rendering results from
      IRON~\cite{Zhang:2022:IRO} (from $500$ colocated training images) on
      the \SceneName{Metallic} scene. Our model is evaluated under colocated point lights. 
      IRON is affected by the interreflections and fails to accurately reconstruct the geometry.}
    \label{fig:comp_iron}
\end{figure*}

%!TEX root = ../../NeuralRelightHint.tex

\newcommand{\nrtfWidth}{{1.45in}}
\newcommand{\nrtfZoomWidth}{{0.5in}}

\begin{figure*}
    \vspace{0.24cm}
    \begin{minipage}{\nrtfWidth} %
      \centering
        \begin{overpic}[width=\textwidth,percent]{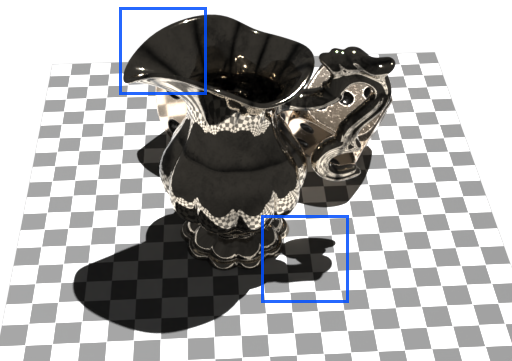}
            \put(3, 78){ \color{black} \small {Reference}}
            \put(3, 71){ \color{black} \small {PSNR | SSIM | LPIPS}}
        \end{overpic}
    \end{minipage} 
    \begin{minipage}{\nrtfZoomWidth}%
        \begin{overpic}[width=\textwidth,percent]{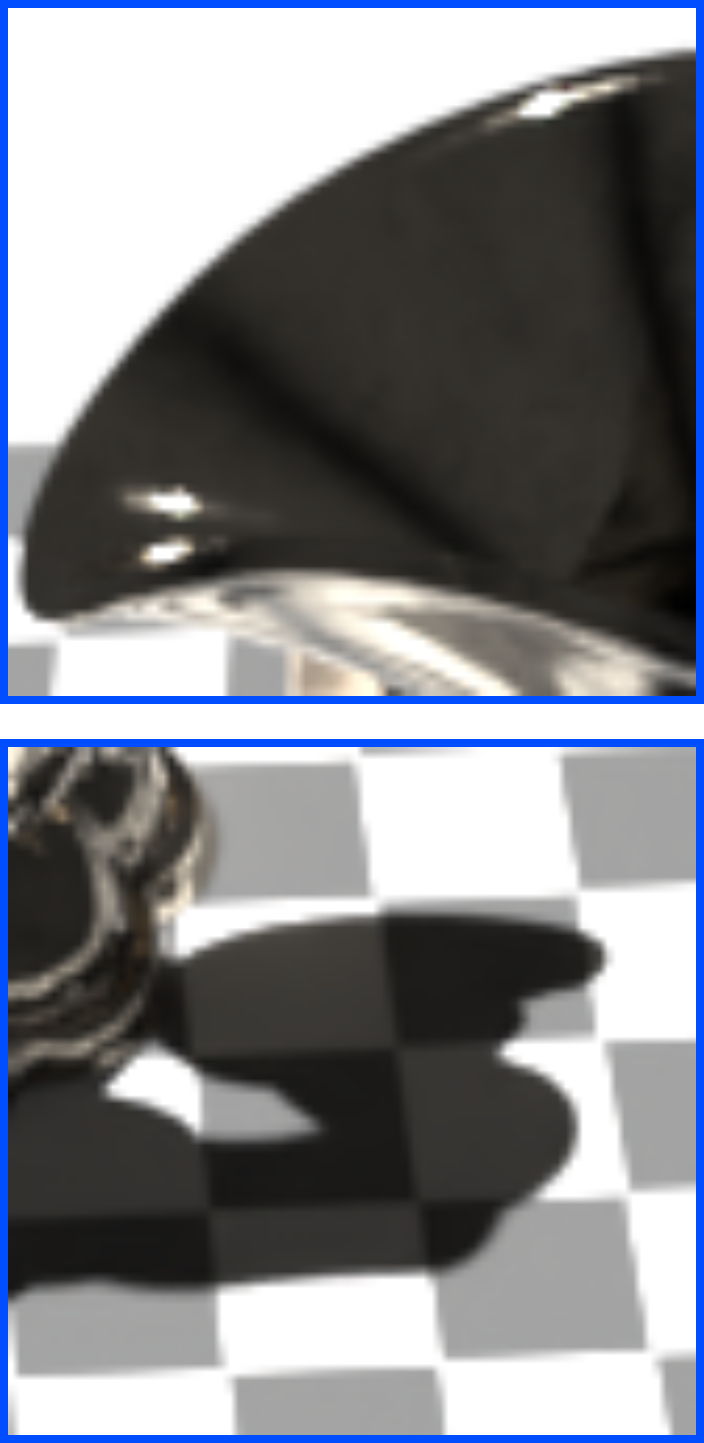}\end{overpic}
    \end{minipage} 
    \begin{minipage}{\nrtfWidth} %
        \begin{overpic}[width=\textwidth,percent]{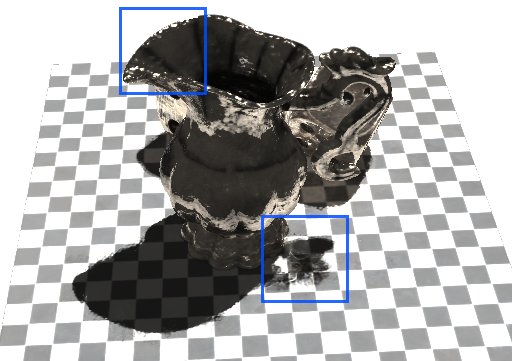}
            \put(3, 78){ \color{black} \small {NRTF}}
            \put(3, 71){ \color{black} \small {22.01 | 0.9008 | 0.1238}}
        \end{overpic}
    \end{minipage}
    \begin{minipage}{\nrtfZoomWidth}%
        \begin{overpic}[width=\textwidth,percent]{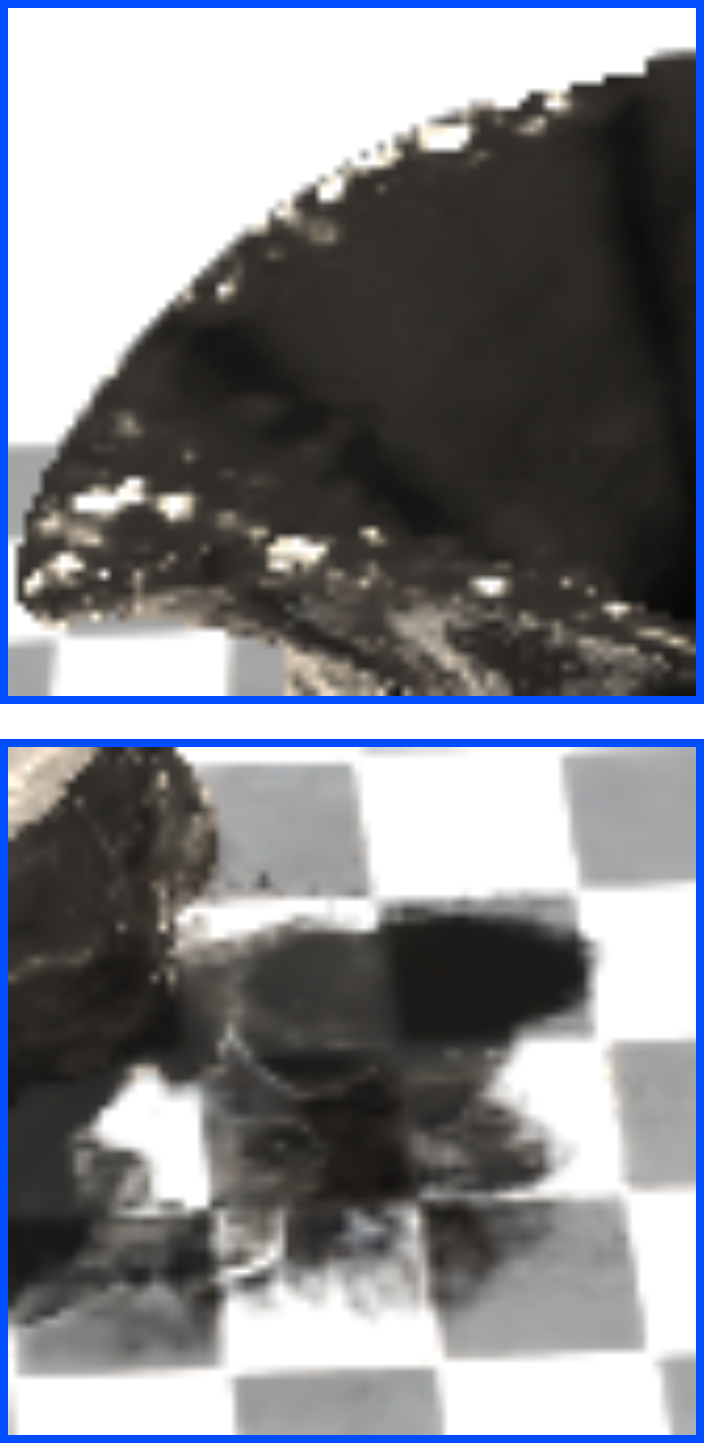}\end{overpic}
    \end{minipage}  
    \begin{minipage}{\nrtfWidth} %
        \begin{overpic}[width=\textwidth,percent]{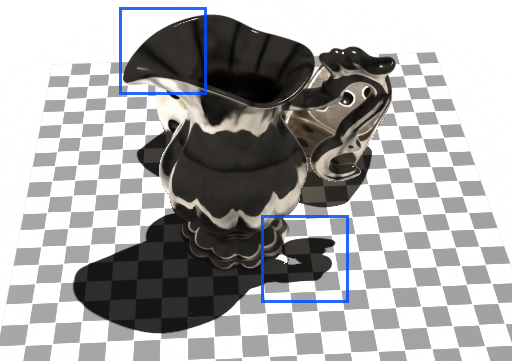}
            \put(3, 78){ \color{black} \small {Ours}}
            \put(3, 71){ \color{black} \small {26.72 | 0.9602 | 0.05351}}
        \end{overpic}
    \end{minipage}  
    \begin{minipage}{\nrtfZoomWidth}%
        \begin{overpic}[width=\textwidth,percent]{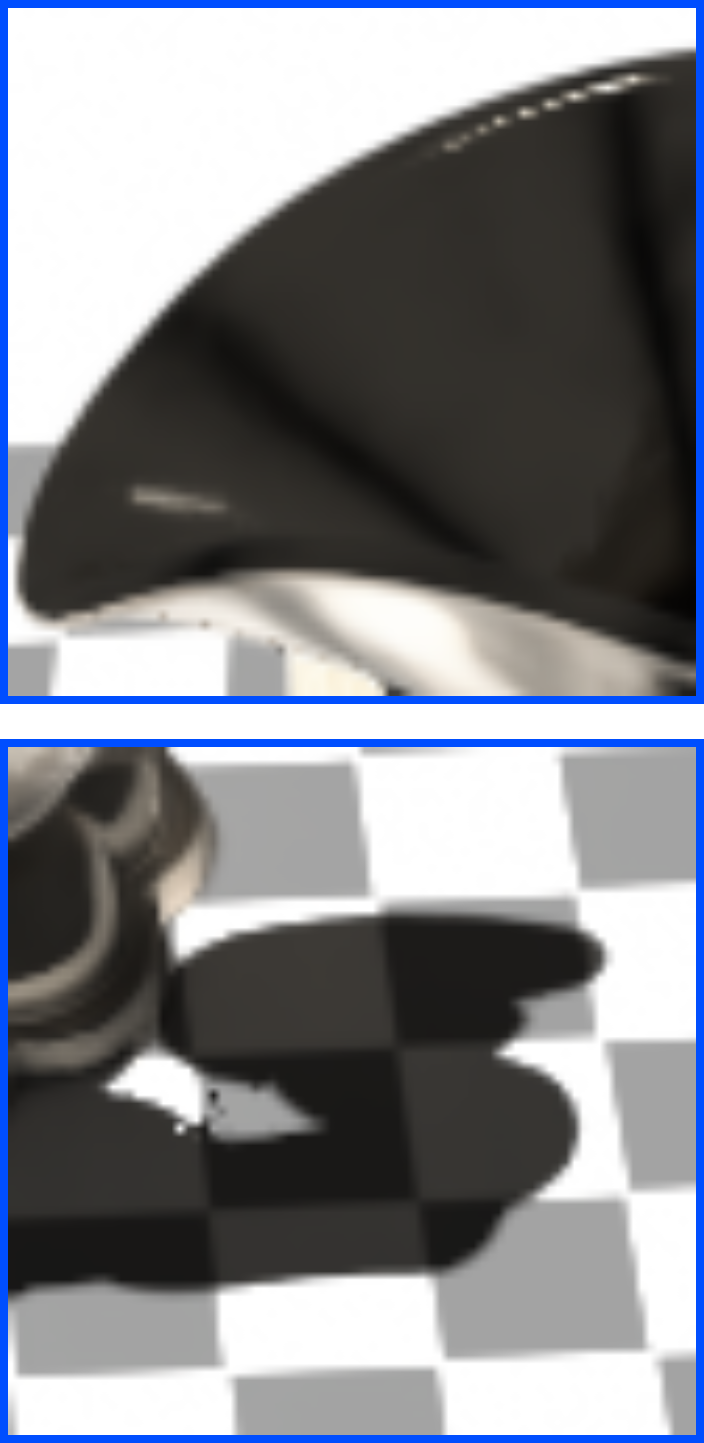}\end{overpic}
    \end{minipage} 
    \caption{A comparison to Neural Radiance Transfer Fields (NRTF) trained on
      $500$ OLAT reference images and reference geometry. To provide a fair
      comparison, we also train our network on the same directional OLAT
      images (without reference geometry) instead of point lighting. NRTF
      struggles to correctly reproduce shadow boundaries and specular
      interreflections (see zoom-ins).}
    \label{fig:comp_nrft}
\end{figure*}

%!TEX root = ../../NeuralRelightHint.tex

\newcommand{\philipFigWidth}{0.21\textwidth}

\begin{figure*}
  \vspace{0.2cm}
  \begin{minipage}{\philipFigWidth} %
    \centering
    \begin{overpic}[width=\textwidth,percent]{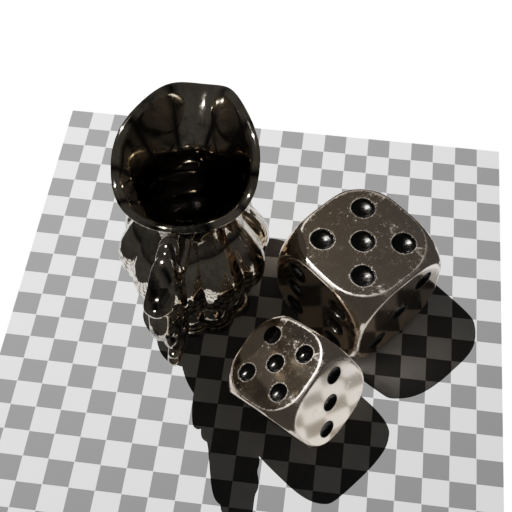}
      \put(3, 101){\color{black} \small {Reference}}
      \put(3, 93){\color{black} \small {PSNR | SSIM | LPIPS}}
    \end{overpic}
  \end{minipage}
  \begin{minipage}{\philipFigWidth} %
    \begin{overpic}[width=\textwidth,percent]{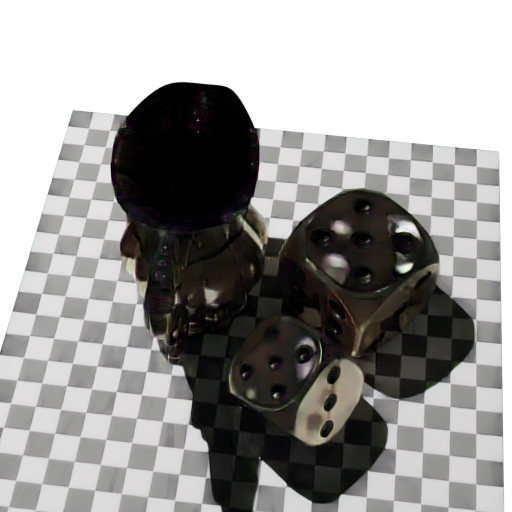}
      \put(3, 101){\color{black} \small {\cite{Philip:2021:FVI}}}
      \put(3, 93){\color{black} \small {w/ reconstructed geometry}}
      \put(3, 85){\color{black} \small {21.29 | 0.8655 | 0.1290}}
    \end{overpic}
  \end{minipage}
  \begin{minipage}{\philipFigWidth} %
    \begin{overpic}[width=\textwidth,percent]{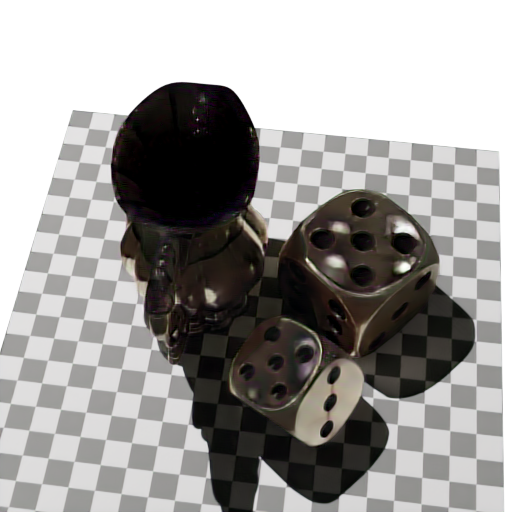}
      \put(3, 101){\color{black} \small {\cite{Philip:2021:FVI}}}
      \put(3, 93){\color{black} \small {w/ reference geometry}}
      \put(3, 85){\color{black} \small {23.22 | 0.8992 | 0.1054}}
    \end{overpic}
  \end{minipage}
  \begin{minipage}{\philipFigWidth} %
    \begin{overpic}[width=\textwidth,percent]{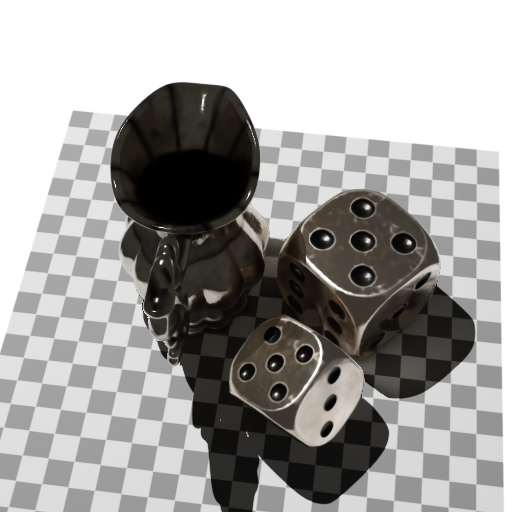}
      \put(3, 101){\color{black}\small  {Ours}}
      \put(3, 93){\color{black}\small  {27.79 | 0.9613 | 0.04873}}
    \end{overpic}
  \end{minipage}
  \caption{Comparison to the pretrained relighting network of
    Philip~\etal~\shortcite{Philip:2021:FVI} on $500$ input images of the
    \SceneName{Metallic} scene rendered with the target lighting. Even under
    these favorable conditions, their method struggles to reproduce the
    correct appearance for novel viewpoints.}
  \label{fig:comp_philip}
\end{figure*}
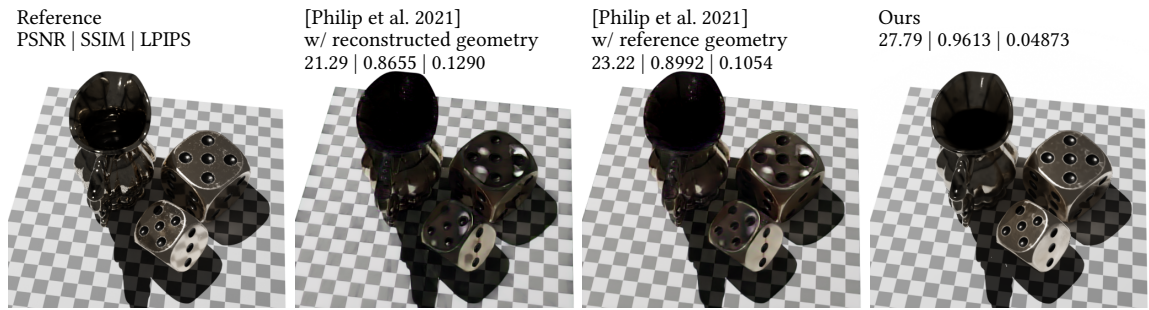

We implemented our neural implicit radiance representation in
PyTorch~\cite{Paszke:2019:PyTorch}.  We train each model for $1,\!000k$ iterations
using the Adam optimizer~\cite{Kingma:2015:Adam} with $\beta_1 = 0.9$ and
$\beta_2 = 0.999$ with $512$ samples per iteration randomly drawn from the
training images. We follow the same warmup and cosine decay learning rate
schedule as in NeuS~\cite{Wang:2021:NLN}. 
Training a single neural implicit radiance representation takes approximate
$20$ hours on four Nvidia V100 GPUs.

We extensively validate the relighting capabilities of our neural implicit
radiance representation on $17$ synthetic and $7$ captured scenes
(including $4$ from~\cite{Gao:2020:DNL}), covering a wide range of different
shapes, materials, and lighting effects.

\paragraph{Synthetic Scenes}
\autoref{fig:baseline} shows relit results of different synthetic scenes. For
each example, we list PSNR, SSIM, and LPIPS~\cite{zhang2018LPIPS} error
statistics computed over $100$ test images different from the $500$ training
images.  Our main test scene contains a vase and two dice; the scene features
a highly concave object (vase) and complex interreflections between the dice.
We include several versions of the main test scene with different material
properties: \SceneName{Diffuse, Metallic, Glossy-Metal, Rough-Metal,
  Anisotropic-Metal, Plastic, Glossy-Plastic, Rough-Plastic} and
\SceneName{Translucent}; note, some versions are only included in the
supplemental material. We also include two versions with modified geometry:
\SceneName{Short-Fur} and \SceneName{Long-Fur} to validate the performance of
our method on shapes with ill-defined geometry.  In addition, we also include
a \SceneName{Fur-Ball} scene which exhibits even longer fur.  To validate the
performance of the shadow hints, we also include scenes with complex shadows:
a \SceneName{Basket} scene containing thin geometric features and a
\SceneName{Layered Woven Ball} which combines complex visibility and strong
interreflections.  In addition to these specially engineered scenes to
systematically probe the capabilities of our method, we also validate our
neural implicit radiance representation on commonly used synthetic scenes
in neural implicit modeling: \SceneName{Hotdog, Lego} and
\SceneName{Drums}~\cite{Mildenhall:2020:NRS}.  Based on the error statistics,
we see that the error correlates with the geometric complexity of the scene
(vase and dice, \SceneName{Hotdog}, and \SceneName{Layered Woven Ball} perform
better than the Fur scenes as well as scenes with small details such as the
\SceneName{Lego} and the \SceneName{Drums} scene), and with the material
properties (highly specular materials such as \SceneName{Metallic} and
\SceneName{Anisotropic-Metal} incur a higher error).  Visually, differences
are most visible in specular reflections and for small geometrical details.

\paragraph{Captured Scenes}
We demonstrate the capabilities of our neural implicit relighting
representation by modeling $3$ new scenes captured with handheld setups 
(\autoref{fig:real}). The \SceneName{Pikachu Statue} scene contains glossy highlights and
significant self-occlusion.  The \SceneName{Cat on Decor} scene showcases the
robustness of our method on real-world objects with ill-defined geometry.  The
\SceneName{Cup and Fabric} scene exhibits translucent materials (cup),
specular reflections of the balls, and anisotropic reflections on the fabric.
%In all cases, our method is able to relight these scenes with full light
%transport under novel point lighting.
We refer to the supplementary material for additional video sequences of these
scenes visualized for rotating camera and light positions.

\paragraph{Comparisons}
\autoref{fig:comp_iron} compares our method to IRON~\cite{Zhang:2022:MII}, an
inverse rendering method that adopts a neural representation for geometry as a
signed distance field. From these results, we can see that IRON fails to
correctly reconstruct the shape and reflections in the presence of strong
interreflections.  In a second comparison (\autoref{fig:comp_nrft}), we
compare our method to Neural Radiance Transfer Fields
(NRTF)~\cite{Lyu:2022:NRT}; we skip the fragile inverse rendering step and
train NRTF with $500$ reference OLAT images and the reference geometry. To
provide a fair comparison, we also train and evaluate our network under the
same directional OLAT images by conditioning the radiance network on light
direction instead of point light position.  From this test we observe that
NRTF struggles to accurately reproduce shadow edges and specular
interreflections, as well as that our method can also be successfully trained
with directional lighting.  \autoref{fig:comp_philip} compares our method to
the pre-trained neural relighting network of
Philip~\etal.~\shortcite{Philip:2021:FVI} on the challenging
\SceneName{Metallic} test scene.  Because multiview stereo
\cite{Schoenberger:2016:SFM} fails for this scene, we input geometry
reconstructed from the NeuS SDF as well as ground truth geometry.  Finally, we
also render the input images under the reference target lighting; our network
is trained without access to the target lighting. Even under these favorable
conditions, the relighting method of Philip~\etal struggles to reproduce the
correct appearance.  Finally, we compare our method to Deferred Neural
Lighting~\cite{Gao:2020:DNL} (using their data and trained model). Our method
is able to achieve similar quality results from $\sim\!\!500$ input images
compared to $\sim \!\! 10,\!000$ input images for Deferred Neural Lighting.
While visually very similar, the overall errors of Deferred Neural Lighting
are slightly lower than with our method.  This is mainly due to differences in
how both methods handle camera calibrations errors. Deferred Neural Lighting
tries to minimize the differences for each frame separately, and thus it can
embed camera calibration errors in the images.  However, this comes at the
cost of temporal ``shimmering'' when calibration is not perfect.  Our method
on the other hand, optimizes the 3D representation, yielding better temporal
stability (and thus requiring less photographs for view interpolation) at the
cost of slightly blurring the images in the presence of camera calibration
errors.
%!TEX root = ../../NeuralRelightHint.tex

\newcommand{\dnlImgWidth}{0.17\textwidth}

\begin{figure*}
\renewcommand{\arraystretch}{0.0}
\addtolength{\tabcolsep}{-5.5pt}
\begin{tabular}{ cccccc }
        \begin{overpic}[height=\dnlImgWidth,percent]{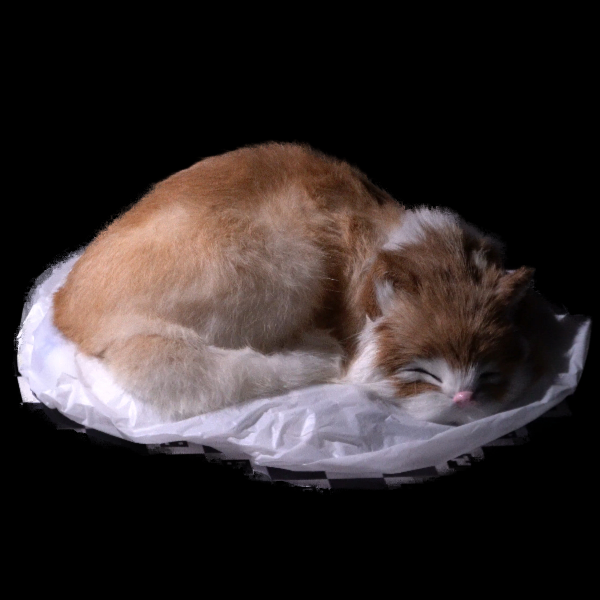}
        \put(3, 90){ \color{white} \small {Reference}}
        \put(3, 83){ \color{white} \footnotesize {PSNR | SSIM | LPIPS}}
        \end{overpic}
        &\begin{overpic}[height=\dnlImgWidth,percent]{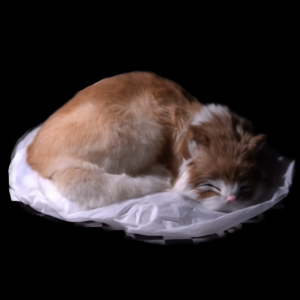}
        \put(3, 90){ \color{white} \small {DNL}}
        \put(3, 83){ \color{white} \footnotesize {39.22 | 0.9932 | 0.0184}}
        \end{overpic}
        &\begin{overpic}[height=\dnlImgWidth,percent]{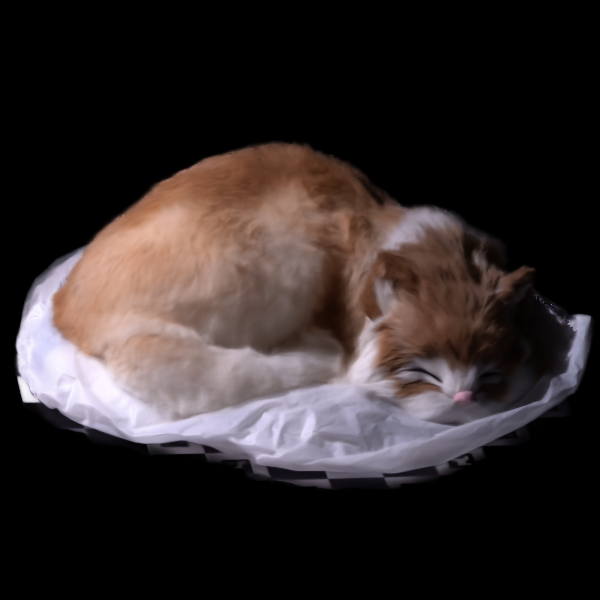}
        \put(3, 90){ \color{white} \small {Ours}}
        \put(3, 83){ \color{white} \footnotesize {36.42 | 0.9856 | 0.0399}}
        \end{overpic}
        &\begin{overpic}[height=\dnlImgWidth,percent]{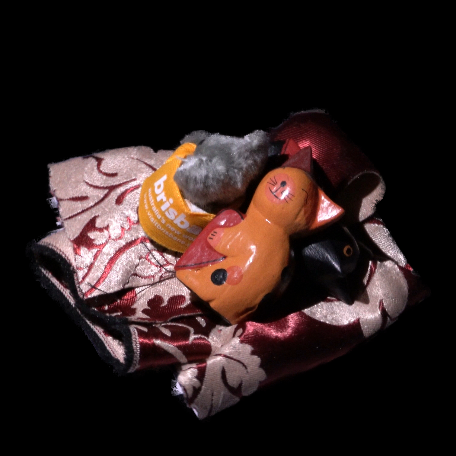}
        \put(3, 90){ \color{white} \small {Reference}}
        \put(3, 83){ \color{white} \footnotesize {PSNR | SSIM | LPIPS}}
        \end{overpic}
        &\begin{overpic}[height=\dnlImgWidth,percent]{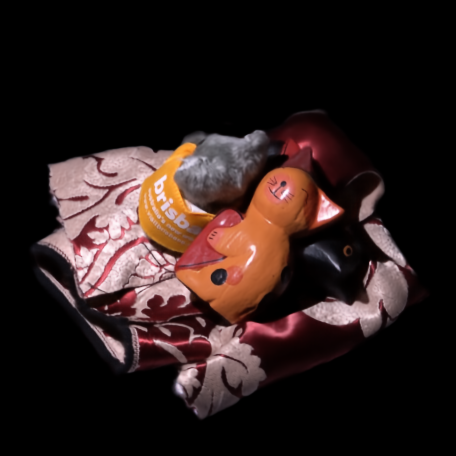}
        \put(3, 90){ \color{white} \small {DNL}}
        \put(3, 83){ \color{white} \footnotesize {34.02 | 0.9763 | 0.0550}}
        \end{overpic}
        &\begin{overpic}[height=\dnlImgWidth,percent]{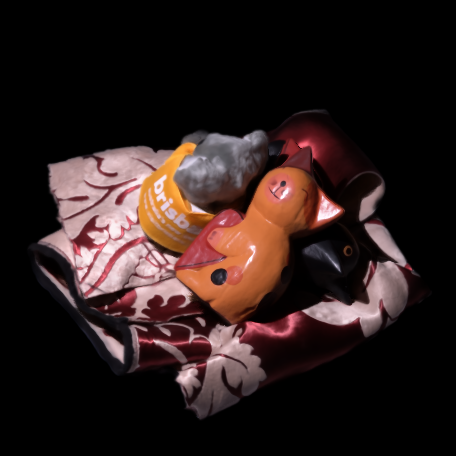}
        \put(3, 90){ \color{white} \small {Ours}}
        \put(3, 83){ \color{white} \footnotesize {32.94 | 0.9708 | 0.0791}}
        \end{overpic}
        \\
        \begin{overpic}[height=\dnlImgWidth,percent]{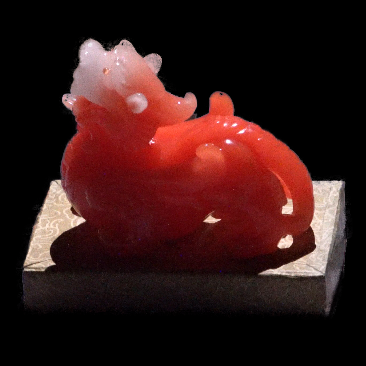}
        \put(3, 93){ \color{white} \footnotesize {PSNR | SSIM | LPIPS}}
        \put(3, 100){ \color{white} \small {Reference}}
        \end{overpic}
        &\begin{overpic}[height=\dnlImgWidth,percent]{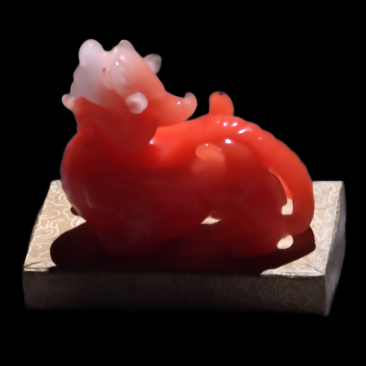}        
        \put(3, 93){ \color{white} \footnotesize {35.36 | 0.9730 | 0.0692}}
        \put(3, 100){ \color{white} \small {DNL}}
        \end{overpic}
        &\begin{overpic}[height=\dnlImgWidth,percent]{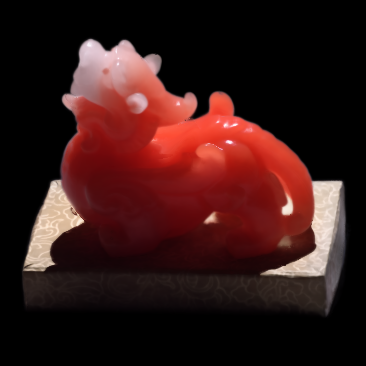}
        \put(3, 93){ \color{white} \footnotesize {33.07 | 0.9695 | 0.0967}}
        \put(3, 100){ \color{white} \small {Ours}}
        \end{overpic}
        &\begin{overpic}[width=\dnlImgWidth,percent]{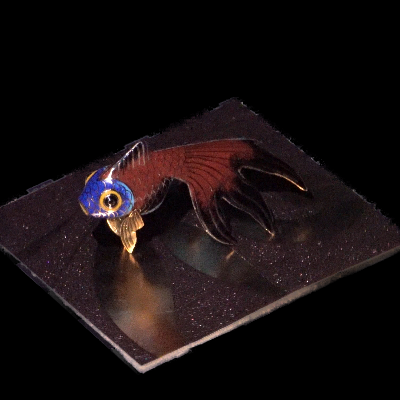}
        \put(3, 93){ \color{white} \footnotesize {PSNR | SSIM | LPIPS}}
        \put(3, 100){ \color{white} \small {Reference}}
        \end{overpic}
        &\begin{overpic}[height=\dnlImgWidth,percent]{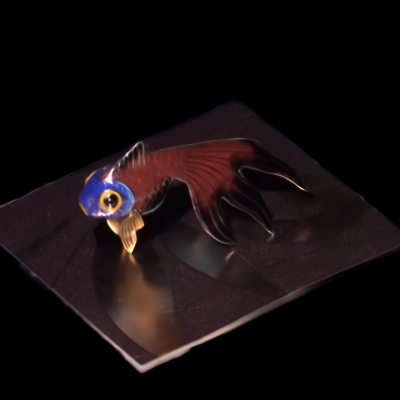}
        \put(3, 93){ \color{white} \footnotesize {32.093 | 0.9469 | 0.1178}}
        \put(3, 100){ \color{white} \small {DNL}}
        \end{overpic}
        &\begin{overpic}[height=\dnlImgWidth,percent]{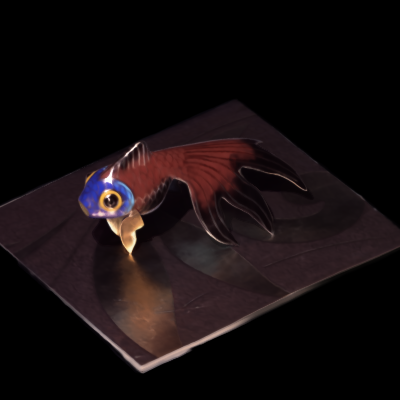}
        \put(3, 93){ \color{white} \footnotesize {30.96 | 0.9445 | 0.1393}}
        \put(3, 100){ \color{white} \small {Ours}}
        \end{overpic}
\end{tabular}
\caption{Comparison with Deferred Neural Lighting~\cite{Gao:2020:DNL}. We
  train our neural implicit radiance representation using only $1/25$th
  ($\sim\!\!500$) randomly selected frames for Gao~\etal\!'s datasets, while
  achieving comparable results.}
  \label{fig:dnl}
\end{figure*}
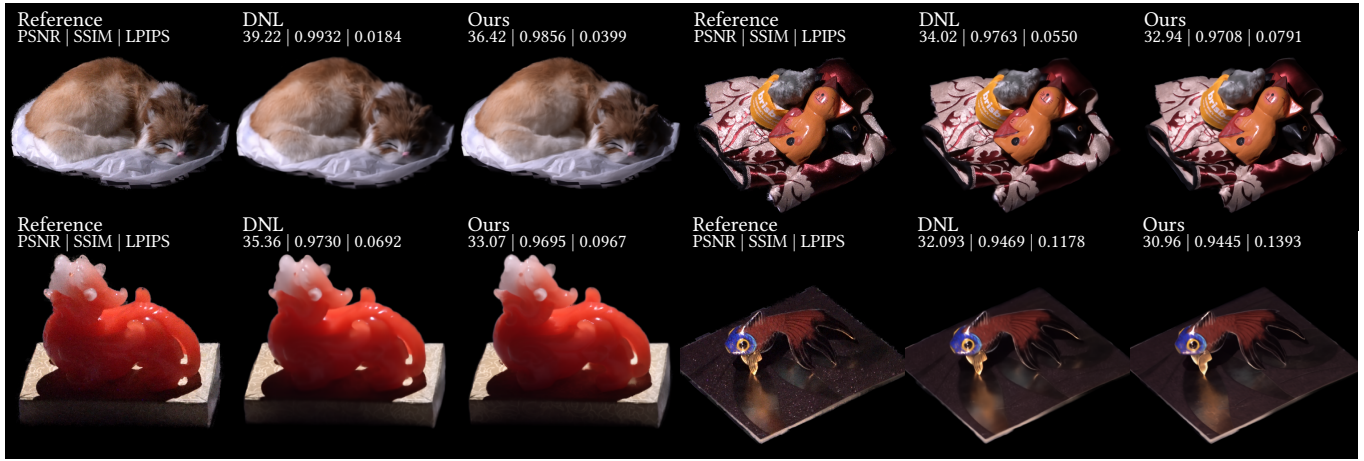

%!TEX root = ../../../NeuralRelightHint.tex
\begin{figure*}
  \begin{minipage}{\textwidth}
    \begin{minipage}{0.03in}
      \hspace{0.03in}
    \end{minipage}
    \begin{minipage}{\textwidth}
      \centering
      \begin{minipage}{\textwidth}
        \centering
        \begin{minipage}{1.3in}
          \centering
          {\small Reference}
        \end{minipage}
        \begin{minipage}{1.3in}
          \centering
          {\small Ours}
        \end{minipage}
        \begin{minipage}{1.3in}
          \centering
          {\small w/o Highlight Hint}
        \end{minipage}
        \begin{minipage}{1.3in}
          \centering
          {\small w/o Shadow Hint}
        \end{minipage}
        \begin{minipage}{1.3in}
          \centering
          {\small w/o Any Hints}
        \end{minipage}
      \end{minipage}
    \end{minipage}
  \end{minipage}

  \begin{minipage}{\textwidth}
    \begin{minipage}{0.03in}
      \centering
      \rotatebox{90}{\small \SceneName{Translucent}}
    \end{minipage}
    \begin{minipage}{\textwidth}
      \centering
      \begin{minipage}{1.3in}
        \includegraphics[width=1.3in]{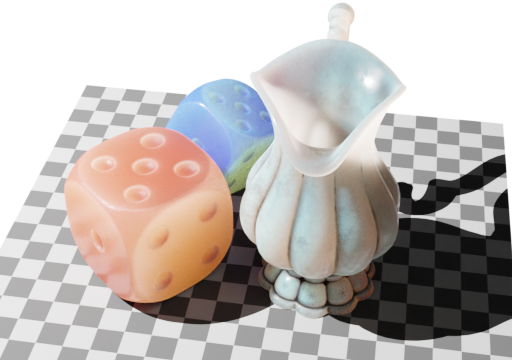}
      \end{minipage}
      \begin{minipage}{1.3in}
        \includegraphics[width=1.3in]{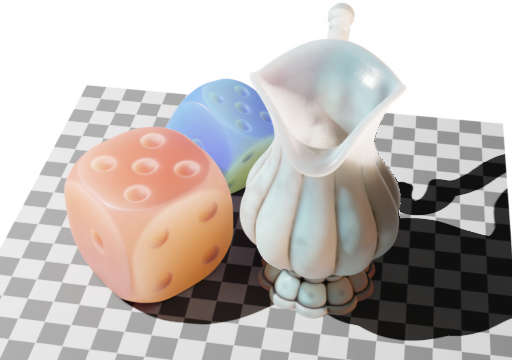}
      \end{minipage}
      \begin{minipage}{1.3in}
        \includegraphics[width=1.3in]{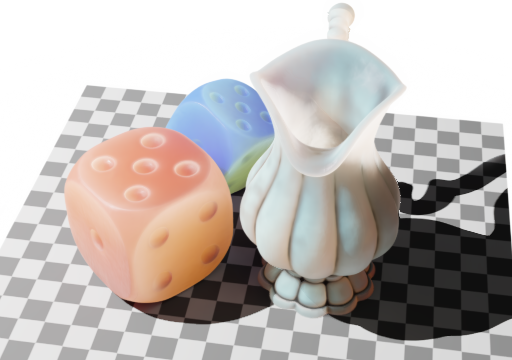}
      \end{minipage}
      \begin{minipage}{1.3in}
        \includegraphics[width=1.3in]{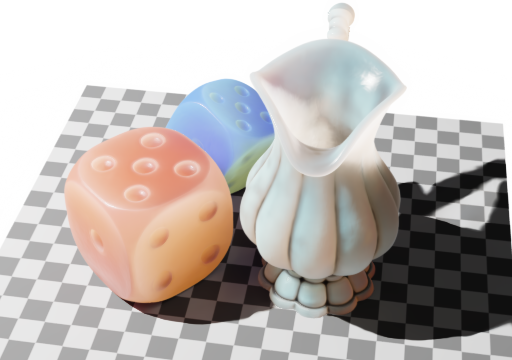}
      \end{minipage}
      \begin{minipage}{1.3in}
        \includegraphics[width=1.3in]{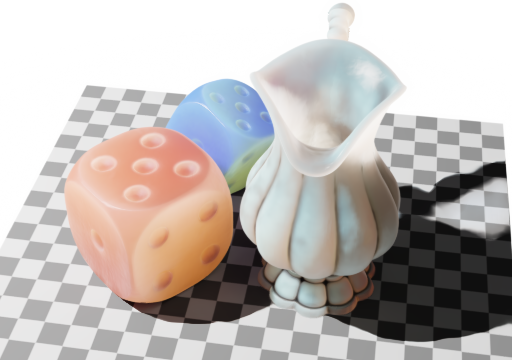}
      \end{minipage}
    \end{minipage}
  \end{minipage}
  \begin{minipage}{\textwidth}
    \begin{minipage}{0.03in}
      \centering
      \rotatebox{90}{\small \SceneName{Layered Woven Ball}}
    \end{minipage}
    \begin{minipage}{\textwidth}
      \centering
      \begin{minipage}{1.3in}
        \includegraphics[width=1.3in]{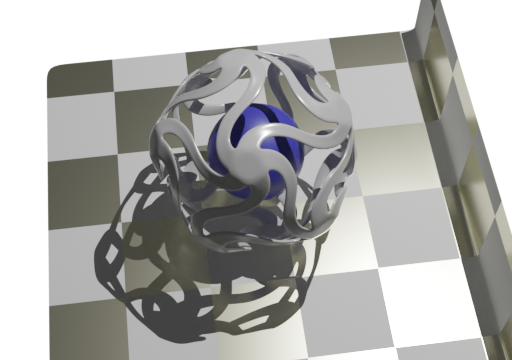}
      \end{minipage}
      \begin{minipage}{1.3in}
        \includegraphics[width=1.3in]{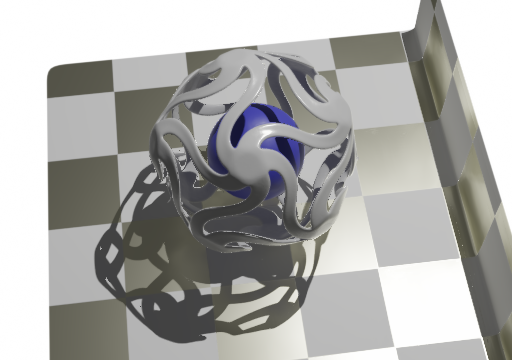}
      \end{minipage}
      \begin{minipage}{1.3in}
        \includegraphics[width=1.3in]{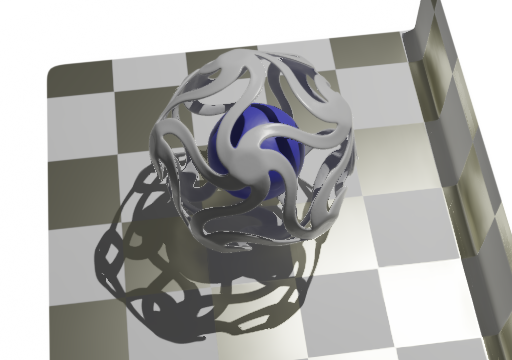}
      \end{minipage}
      \begin{minipage}{1.3in}
        \includegraphics[width=1.3in]{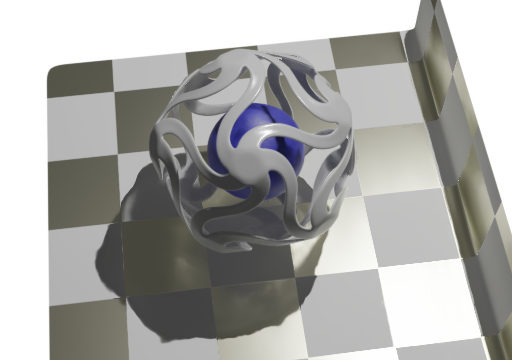}
      \end{minipage}
      \begin{minipage}{1.3in}
        \includegraphics[width=1.3in]{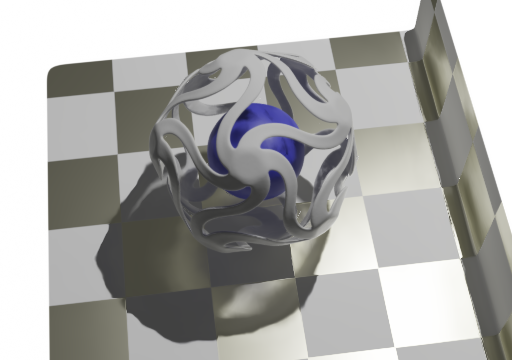}
      \end{minipage}
    \end{minipage}
  \end{minipage}
  \caption{Impact of shadow and highlight hints; without the hints the
    network fails to accurately reproduce the desired effect.}
  \label{fig:hints}
\end{figure*}

%!TEX root = ../../../NeuralRelightHint.tex
\begin{figure*}
\newcommand{\shadowFigWidth}{{1.3in}}
\renewcommand{\arraystretch}{0.3}
\addtolength{\tabcolsep}{-5.5pt}
\begin{tabular}{ cccc }
        \small Reference & \small $16$ shadow rays & \small $1$ shadow ray (Ours) & \small NeRF $1$ shadow ray\\
        \footnotesize PSNR | SSIM | LPIPS 
        & \footnotesize 28.22 | 0.9667 | 0.0365 
        & \footnotesize 26.84 | 0.9586 | 0.0411 
        & \footnotesize 23.71 | 0.9160 | 0.0733\\
        \includegraphics[width=\shadowFigWidth]{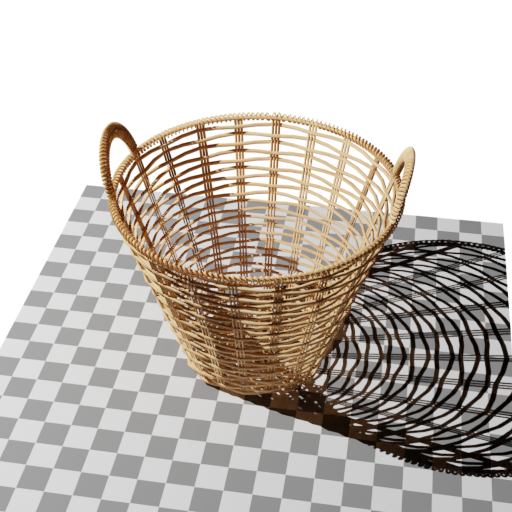}
        &\includegraphics[width=\shadowFigWidth]{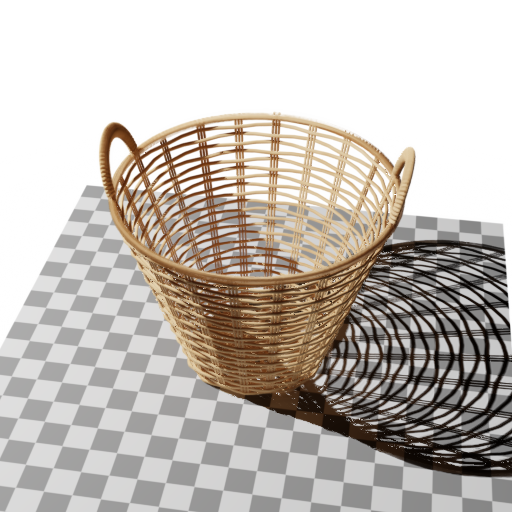}
        &\includegraphics[width=\shadowFigWidth]{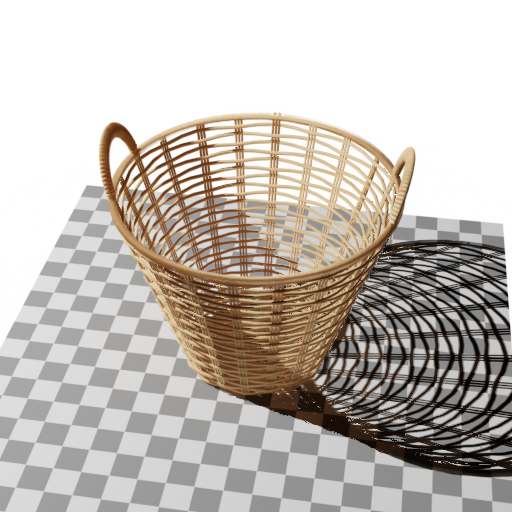}
        &\includegraphics[width=\shadowFigWidth]{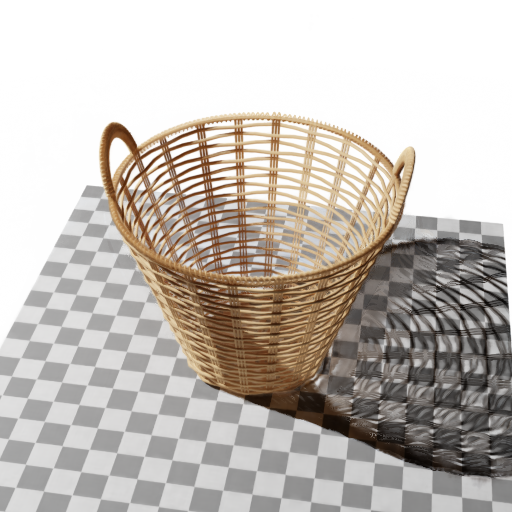}
\end{tabular}
\caption{Impact of the number of shadow rays and the underlying implicit shape
  representation demonstrated on the \SceneName{Basket} scene. Using $16$
  shadow rays only provides marginal improvements at the cost of significant
  computation overhead.  Using NeRF as the basis for the neural implicit shape
  yields degraded shadow quality due to depth biases.  }
    \label{fig:shadow}
\end{figure*}

%!TEX root = ../../../NeuralRelightHint.tex
%\begin{minipage}{0.46\textwidth}
\begin{figure*}
  \newcommand{\numimgFigWidth}{0.18\textwidth}
  \renewcommand{\arraystretch}{0.3}
  \addtolength{\tabcolsep}{-5.5pt}
  \begin{tabular}{ ccccc }
    Reference & $50$ inputs & $100$ inputs & $250$ inputs & $500$ inputs \\
    \includegraphics[width=\numimgFigWidth]{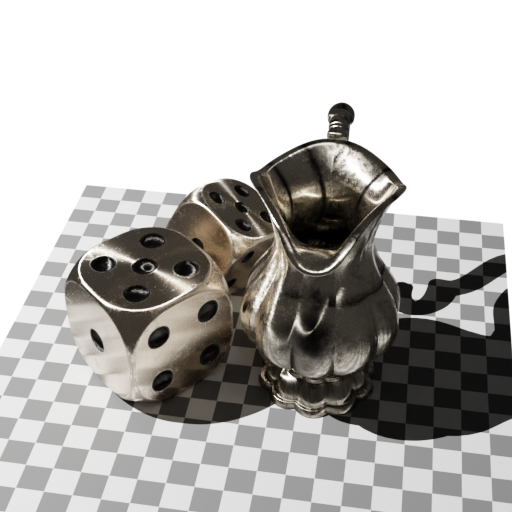}
  & \includegraphics[width=\numimgFigWidth]{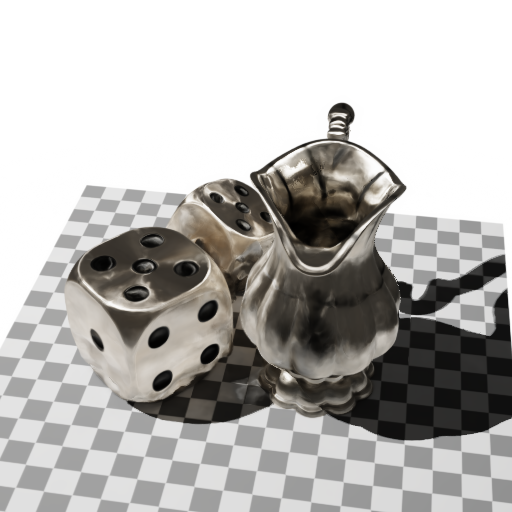}
  & \includegraphics[width=\numimgFigWidth]{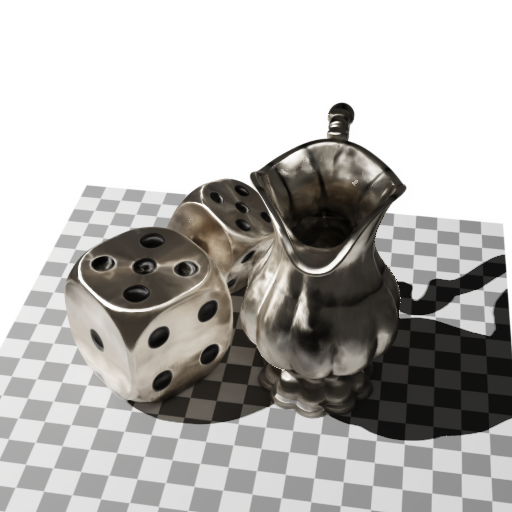}
  & \includegraphics[width=\numimgFigWidth]{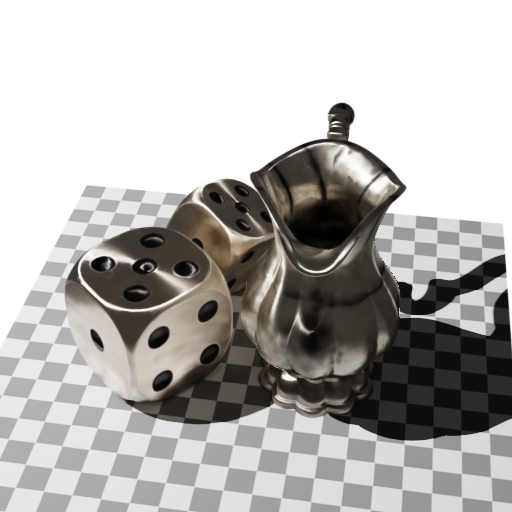}
  & \includegraphics[width=\numimgFigWidth]{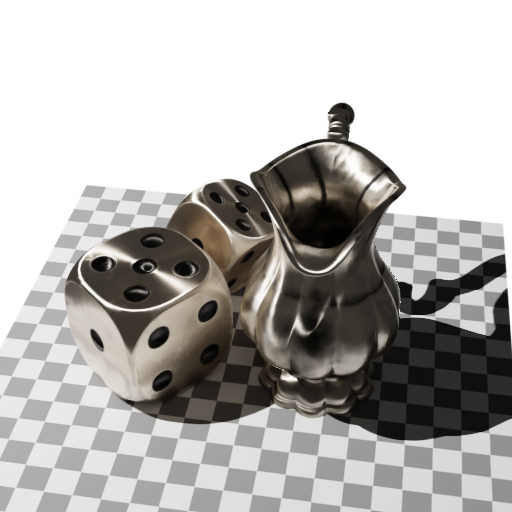}
  \end{tabular}
  \caption{Impact of the number of captured training images. Increasing
    the number of training images improves the quality.  The quality
    degrades significantly when the number of images is less than $250$. }
  \label{fig:input}
  %\end{minipage}
\end{figure*}

%!TEX root = ../../NeuralRelightHint.tex

\newcommand{\camOptFigWidth}{0.23\textwidth}
\newcommand{\zoomWidth}{{0.5in}}

\begin{figure*}
  \begin{minipage}{\camOptFigWidth} %
    \centering
    \begin{overpic}[width=\textwidth,percent]{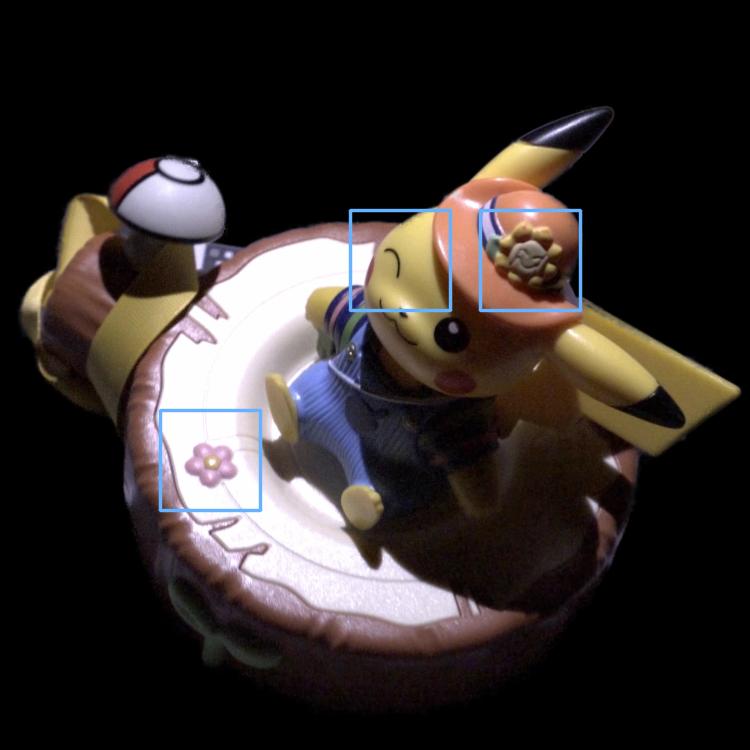}
      \put(3, 91){ \color{white} \small {Reference}}
      \put(3, 83){ \color{white} \small {PSNR | SSIM | LPIPS}}
    \end{overpic}
  \end{minipage}
  \begin{minipage}{\zoomWidth}%
    \begin{overpic}[width=\textwidth,percent]{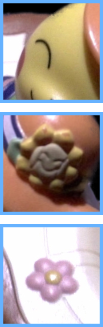}\end{overpic}
  \end{minipage}
  \begin{minipage}{\camOptFigWidth} %
    \begin{overpic}[width=\textwidth,percent]{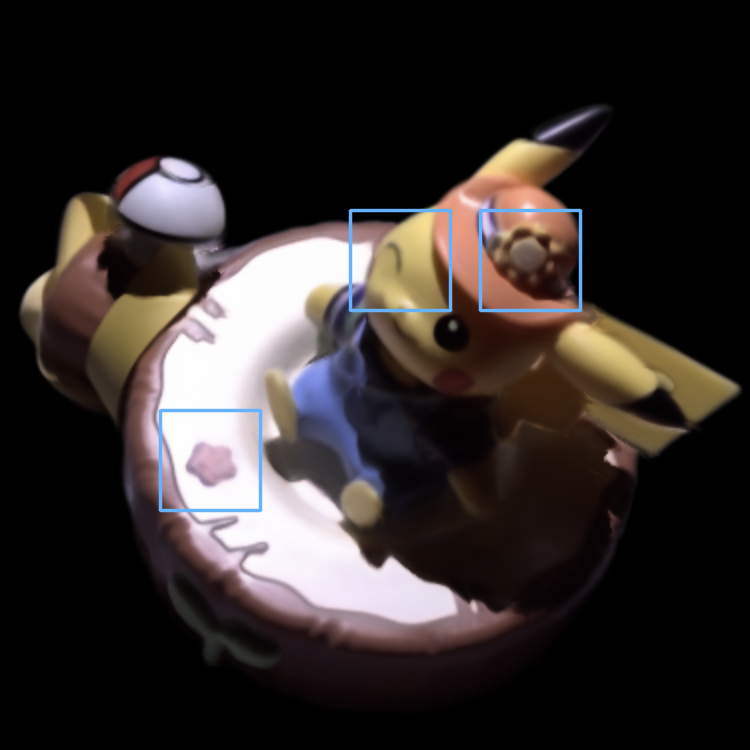}
      \put(3, 91){ \color{white} \small {w/o Viewpoint Optimization}}
      \put(3, 83){ \color{white} \small {31.43 | 0.9803 | 0.0375}}
    \end{overpic}
  \end{minipage}
  \begin{minipage}{\zoomWidth}%
    \begin{overpic}[width=\textwidth,percent]{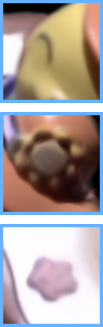}\end{overpic}
  \end{minipage}
  \begin{minipage}{\camOptFigWidth} %
    \begin{overpic}[width=\textwidth,percent]{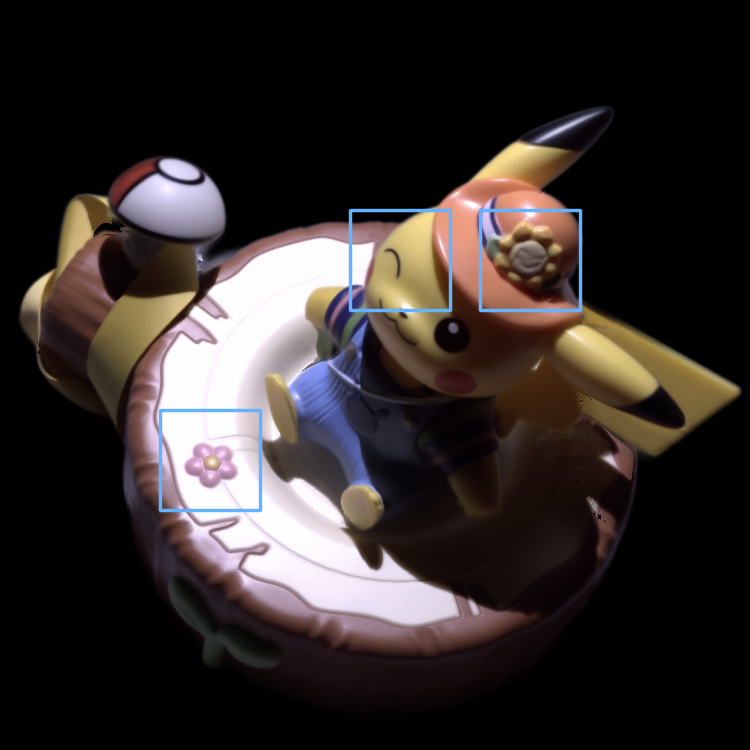}
      \put(3, 91){ \color{white}\small  {w/ Viewpoint Optimization}}
      \put(3, 83){ \color{white} \small {35.08 | 0.9877 | 0.0.359}}
    \end{overpic}
  \end{minipage}
  \begin{minipage}{\zoomWidth}%
    \begin{overpic}[width=\textwidth,percent]{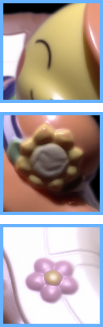}\end{overpic}
  \end{minipage}
  \caption{Effectiveness of Viewpoint Optimization. Using viewpoint optimization greatly enhances the image quality in terms of sharpness and detail.}
  \label{fig:cam_opt}
\end{figure*}

\section{Ablation Studies}
\label{sec:ablation}

We perform several ablation experiments (visual and quantitative) on the
synthetic datasets to evaluate the impact of each of the components that
comprise our neural implicit radiance representation.

% ablation result table
\begin{table}[]
    \caption{Ablation results on synthetic scenes}
    \vspace{-0.2cm}
    \label{tab:ablation_results}
    \begin{tabular}{l|ccc}
    \hline
    Ablation Variant             & PSNR $\uparrow$ & SSIM $\uparrow$ & LPIPS $\downarrow$ \\ \hline
    \textbf{Full hints}          & 32.02           & 0.9727          & 0.0401             \\
    w/o highlight hint           & 31.96           & 0.9724          & 0.0407             \\
    w/o shadow hint              & 27.67           & 0.9572          & 0.0610             \\
    w/o any hints                & 27.54           & 0.9568          & 0.0620             \\ \hline
    1 basis material             & 31.54           & 0.9707          & 0.0428             \\
    2 basis materials            & 31.54           & 0.9707          & 0.0429             \\
    \textbf{4 basis materials}   & 32.02           & 0.9727          & 0.0401             \\
    8 basis materials            & 31.98           & 0.9726          & 0.0401             \\ \hline
    50 training images           & 24.29           & 0.9335          & 0.0706             \\
    100 training images          & 27.96           & 0.9572          & 0.0520             \\
    250 training images          & 30.36           & 0.9666          & 0.0456             \\
    \textbf{500 training images} & 32.02           & 0.9727          & 0.0401             \\ \hline
    \end{tabular}
    \vspace{-0.2cm}
\end{table}

% cam opt result table
\begin{table}[]
    \caption{Ablation results of viewpoint optimization on real captured scenes}
    \vspace{-0.2cm}
    \label{tab:ablation_cam_opt}
    \begin{tabular}{l|ccc}
    \hline
    Ablation Variant             & PSNR $\uparrow$ & SSIM $\uparrow$ & LPIPS $\downarrow$ \\ \hline
    \textbf{w/ viewpoint optimization}          & 34.72           & 0.9762          & 0.0695             \\
    w/o viewpoint optimization                & 33.62           & 0.9719          & 0.0794             \\ \hline
    \end{tabular}
    \vspace{-0.2cm}
\end{table}

\paragraph{Shadow and Highlight Hints}
A key contribution is the inclusion of shadow and highlight hints in the
relightable radiance MLP. \autoref{fig:hints} shows the impact of training
without the shadow hint, the highlight hint, or both.  Without shadow hints
the method fails to correctly reproduce sharp shadow boundaries on the ground
plane.  This lack of sharp shadows is also reflected in the quantitative
errors summarized in~\autoref{tab:ablation_results}.  Including the highlight
hints yield a better highlight reproduction, \eg\!, in the mouth of the vase.

% ablation on hints figure
%\input{src/results/ablations/hint_figure.tex}

\paragraph{Impact of the Number of Shadow Rays}
We currently only use a single shadow ray to compute the shadow hint.
However, we can also shoot multiple shadow rays (by importance sampling points
along the view ray) and provide a more accurate hint to the radiance
network.  \autoref{fig:shadow} shows the results of a radiance network
trained with $16$ shadow rays. While providing a more accurate shadow hint,
there is marginal benefit at a greatly increased computational cost,
justifying our choice of a single shadow ray for computing the shadow hint.

\paragraph{NeuS vs. NeRF Density MLP}
While the relightable radiance MLP learns how much to trust the shadow hint (worst case
it can completely ignore unreliable hints), the radiance MLP can in general
not reintroduce high-frequency details if it is not included in the shadow
hints. To obtain a good shadow hint, an accurate depth estimate of the mean
depth along the view ray is needed.  Wang~\etal~\shortcite{Wang:2021:NLN}
noted that NeRF produces a biased depth estimate, and they introduced NeuS to
address this problem.  Replacing NeuS by NeRF for the density network
(\autoref{fig:shadow}) leads to poor shadow reproduction due to the adverse
impact of the biased depth estimates on the shadow hints.

\paragraph{Impact of the number of Basis Materials for the Highlight Hints}
\autoref{tab:ablation_results} shows the results of using $1, 2, 4$ and $8$
basis materials for computing the highlight hints. Additional highlights hints
improve the results up to a point; when too many hints are provided erroneous
correlations can increase the overall error.  $4$ basis materials strike a
good balance between computational cost, network complexity, and quality.

\paragraph{Impact of Number of Training Images}
\autoref{fig:input} and \autoref{tab:ablation_results} demonstrate the effect
of varying the number of input images from $50$, $100$, $250$ to $500$. As
expected, more training images improve the results, and with increasing number
of images, the increase in improvement diminishes.  With $250$ images we
already achieve plausible relit results.  Decreasing the number of training
images further introduces noticeable appearance differences.

\paragraph{Effectiveness of Viewpoint Optimization}
\autoref{fig:cam_opt} and \autoref{tab:ablation_cam_opt} demonstrate the
effectiveness of viewpoint optimization on real captured scenes. While the
improvement in quantitative errors is limited, visually we can see that
viewpoint optimization significantly enhances reconstruction quality with
increased sharpness and better preservation of finer details.

\section{Limitations}
\label{sec:limitations}
While our neural implicit radiance representation greatly reduces the number
of required input images for relighting scenes with complex shape and
materials, it is not without limitations.  Currently we provide shadow and
highlight hints to help the relightable radiance MLP model high frequency
light transport effects.  However, other high frequency effects exist.  In
particular highly specular surfaces that reflect other parts of the scene pose
a challenge to the radiance network.  Na\"ive inclusion of \emph{'reflection
  hints'} and/or reparameterizations~\cite{Verbin:2022:RSV} fail to help the
network, mainly due to the reduced accuracy of the surface normals (needed to
predict the reflected direction) for sharp specular materials.  Resolving this
limitation is a key challenge for future research in neural implicit modeling
for image-based relighting.

\section{Conclusion}
In this paper we presented a novel neural implicit radiance representation for
free viewpoint relighting from a small set of unstructured photographs.  Our
representation consists of two MLPs: one for modeling the SDF (analogous to
NeuS) and a second MLP for modeling the local and indirect radiance at each
point. Key to our method is the inclusion of shadow and highlight hints to aid
the relightable radiance MLP to model high frequency light transport effects.
Our method is able to produce relit results from just $\sim\!500$
photographs of the scene; a saving of one to two order of magnitude compared
to prior work with similar capabilities.

\begin{acks}
Pieter Peers was supported in part by NSF
grant IIS-1909028. Chong Zeng and Hongzhi Wu were partially supported by NSF China (62022072 \& 62227806), Zhejiang Provincial Key R\&D Program (2022C01057) and the XPLORER PRIZE.
\end{acks}

\bibliographystyle{ACM-Reference-Format}
\bibliography{src/reference}

\clearpage

%%%%%%%%%%%%%%%%%%%%%%%%%%%%%%%%%%%%%%%%%%%%%%%%%%%%%%%%%%%%%%%%%%%
\end{document}

\title[Relighting Neural Radiance Fields with Shadow and Highlight Hints]{Relighting Neural Radiance Fields with \\ Shadow and Highlight Hints}

\author{Chong Zeng}
\authornote{Work done during internship at Microsoft Research Asia.}
\affiliation{%
  \institution{State Key Lab of CAD and CG, Zhejiang University}
  %\institution{Microsoft Research Asia}
  \city{Hangzhou}
  \country{China}
}
\orcid{0009-0004-6373-6848}
\email{chongzeng2000@gmail.com}

\author{Guojun Chen}
\affiliation{%
  \institution{Microsoft Research Asia}
  \city{Beijing}
  \country{China}
}
\orcid{0000-0002-3207-6283}
\email{guoch@microsoft.com}

\author{Yue Dong}
\affiliation{%
  \institution{Microsoft Research Asia}
  \city{Beijing}
  \country{China}
}
\orcid{0000-0003-0362-337X}
\email{yuedong@microsoft.com}

\author{Pieter Peers}
\affiliation{%
  \institution{College of William \& Mary}
  \city{Williamsburg}
  \country{USA}
}
\orcid{0000-0001-7621-9808}
\email{ppeers@siggraph.org}

\author{Hongzhi Wu}
\affiliation{%
  \institution{State Key Lab of CAD and CG, Zhejiang University}
  \city{Hangzhou}
  \country{China}
}
\orcid{0000-0002-4404-2275}
\email{hwu@acm.org}

\author{Xin Tong}
\affiliation{%
  \institution{Microsoft Research Asia}
  \city{Beijing}
  \country{China}
}
\orcid{0000-0001-8788-2453}
\email{xtong@microsoft.com}
%\authorsaddresses{}

\renewcommand{\shortauthors}{Zeng et al.}
%\author{Submission ID: 113}
%!TEX root = ../../../NeuralRelightHint_supplementary.tex

\newcommand{\synImgWidth}{0.16\textwidth}

\begin{teaserfigure}
\renewcommand{\arraystretch}{0.3}
\addtolength{\tabcolsep}{-5.5pt}
\begin{tabular}{ cccccc }
        \begin{overpic}[width=\synImgWidth,percent]{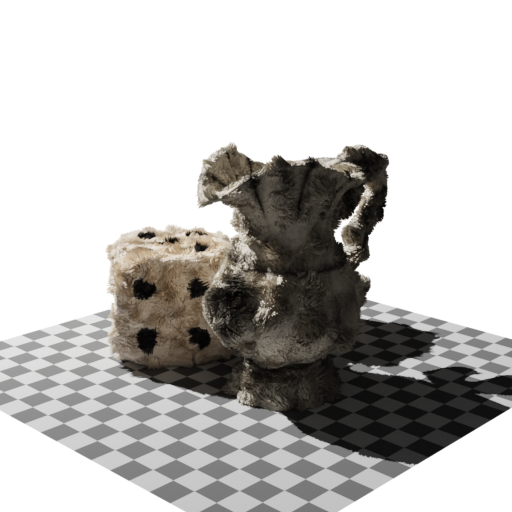}\end{overpic}
        &\begin{overpic}[width=\synImgWidth,percent]{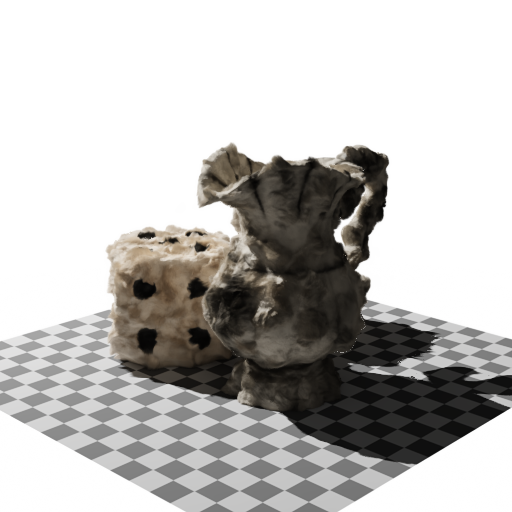}\end{overpic}
        &\begin{overpic}[width=\synImgWidth,percent]{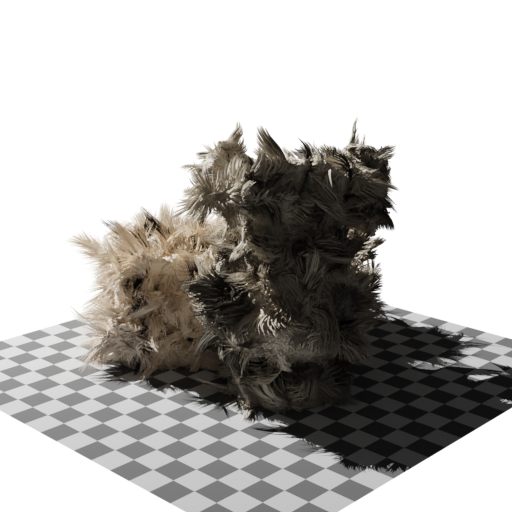}\end{overpic}
        &\begin{overpic}[width=\synImgWidth,percent]{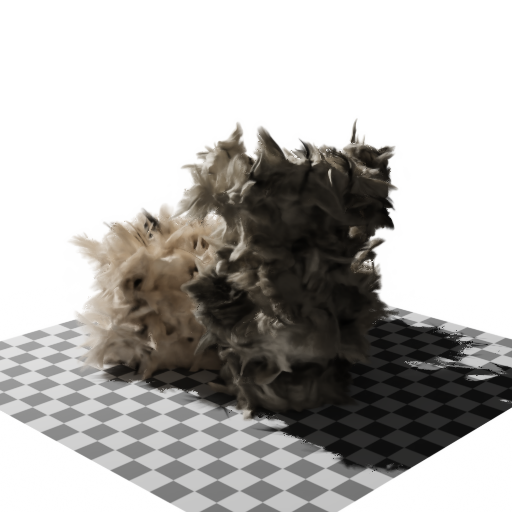}\end{overpic}
        &\begin{overpic}[width=\synImgWidth,percent]{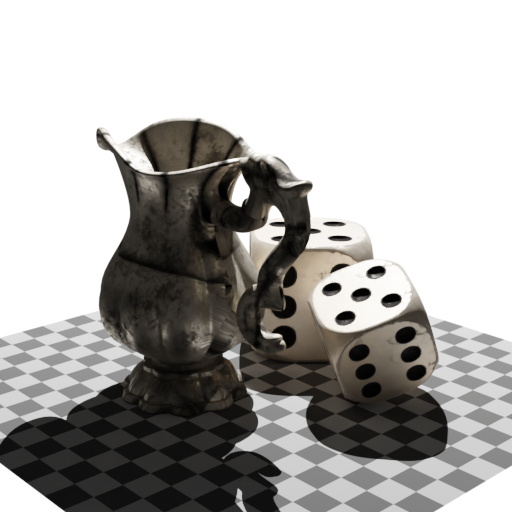}\end{overpic}
        &\begin{overpic}[width=\synImgWidth,percent]{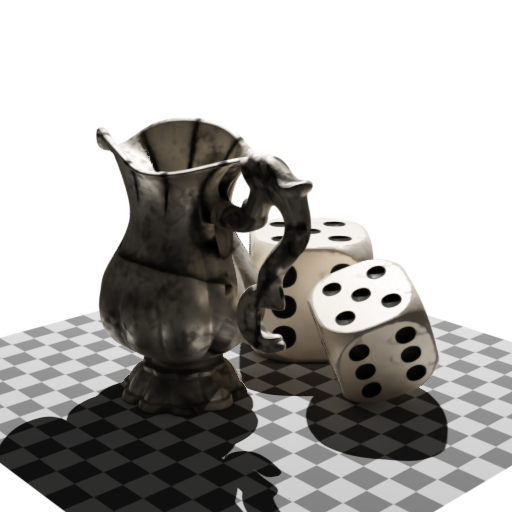}\end{overpic}
        \\
        \multicolumn{2}{c}{\small {\SceneName{Short Fur}: 29.70 | 0.9598 | 0.0823}}
        &\multicolumn{2}{c}{\small {\SceneName{Long Fur}: 25.53 | 0.9060 | 0.1345}}
        &\multicolumn{2}{c}{\small {\SceneName{Rough-Metal}: 35.75 | 0.9908 | 0.0176}} \\
        \begin{overpic}[width=\synImgWidth,percent]{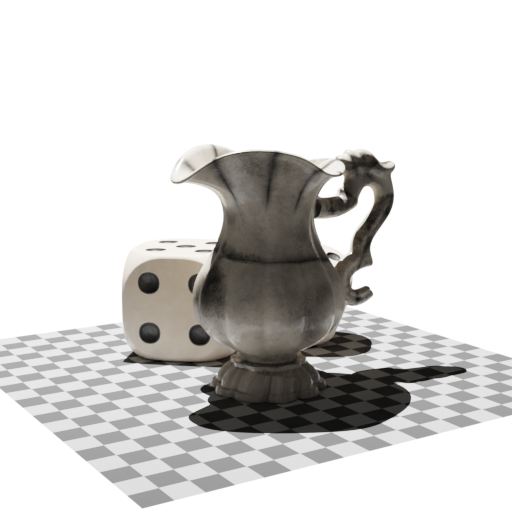}\end{overpic}
        &\begin{overpic}[width=\synImgWidth,percent]{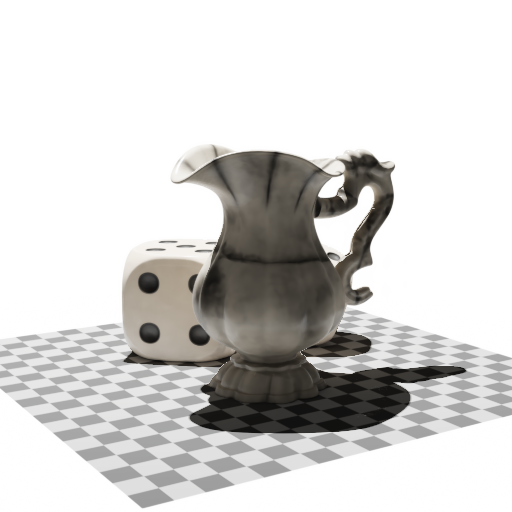}\end{overpic}
        &\begin{overpic}[width=\synImgWidth,percent]{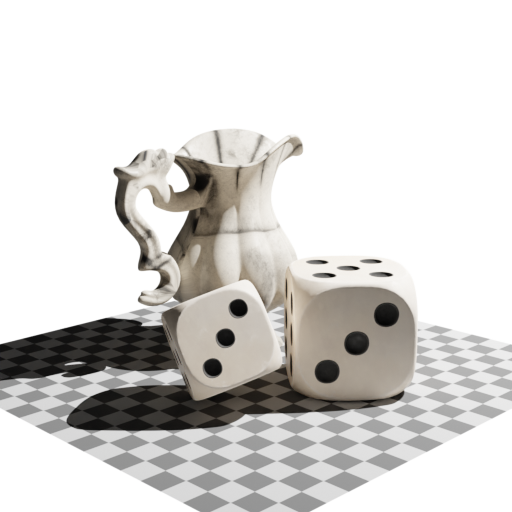}\end{overpic}
        &\begin{overpic}[width=\synImgWidth,percent]{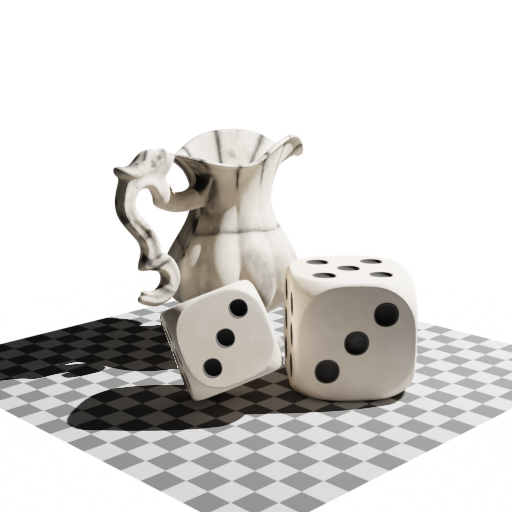}\end{overpic}
        &\begin{overpic}[width=\synImgWidth,percent]{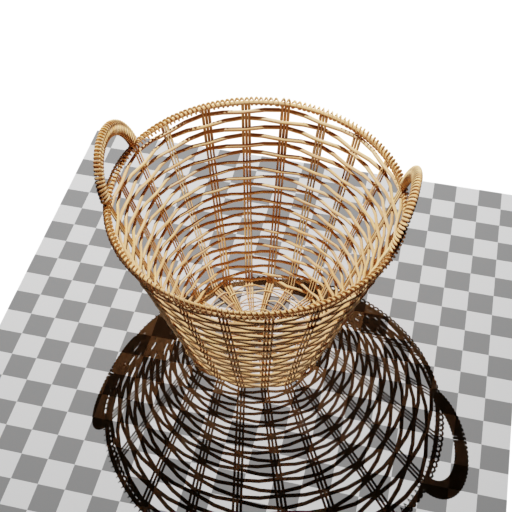}\end{overpic}
        &\begin{overpic}[width=\synImgWidth,percent]{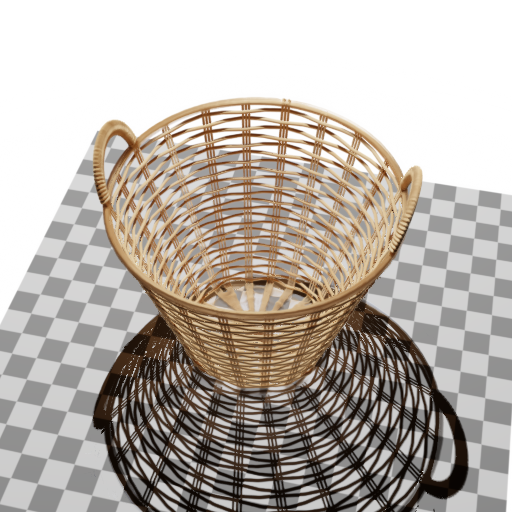}\end{overpic}
        \\
        \multicolumn{2}{c}{\small {\SceneName{Glossy-Plastic}: 36.16 | 0.9920 | 0.0155}} 
        &\multicolumn{2}{c}{\small {\SceneName{Rough-Plastic}: 37.41 | 0.9945 | 0.0132 }}
        &\multicolumn{2}{c}{\small {\SceneName{Basket}: 26.84 | 0.9586 | 0.0411  }}
\end{tabular}
\caption{Qualitative comparison between additional synthetic scenes relit
  (right) for a novel viewpoint and novel lighting direction (not part of the
  training data) and a rendered reference image (left). For each example we
  list average PSNR, SSIM, and LPIPS computed over a uniform sampling of view
  and light positions.}
  \label{fig:baseline}
\end{teaserfigure}

\maketitle

%!TEX root = ../NeuralRelightHint_supplementary.tex

\section{Additional results}
\autoref{fig:baseline} shows additional synthetic results to further test our
method on scenes with different material properties.  The \SceneName{Basket}
scene is included in the ablation study figures, but not listed in Figure 3
(of the main paper); we include it here for completeness.

%!TEX root = ../NeuralRelightHint.tex
\begin{figure}
  \includegraphics[width=\textwidth,width=0.45\textwidth]{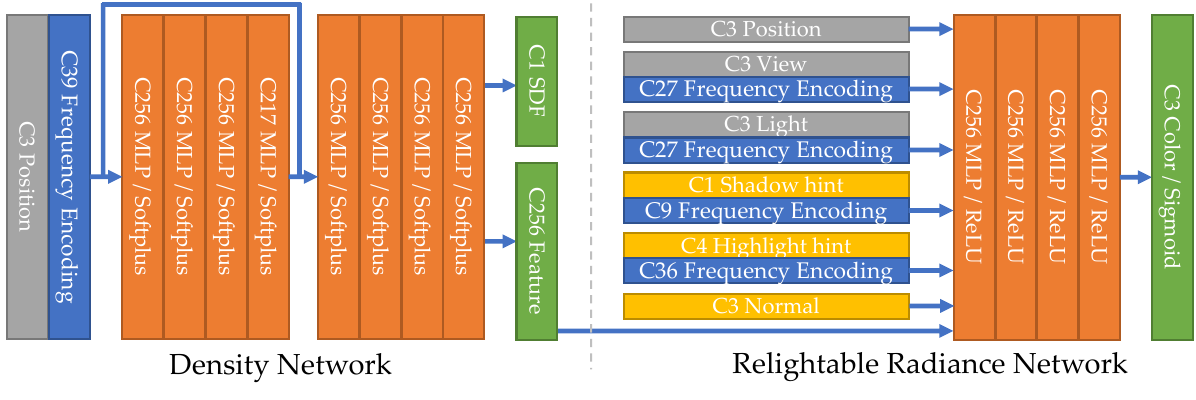}
  \caption{Detailed network architecture of the density and relightable radiance network. 
  The number of output channels and activations are also marked.}
  \label{fig:networks}
\end{figure}

\section{Network Architexture Details}
We follow exactly the same architecture as NeuS~\cite{Wang:2021:NLN} for the
density MLP: $8$ hidden layers with $256$ nodes using a Softplus activation
and a skip connection between the input and the $4$th layer.  The input (i.e.,
current position along a ray) is augmented using a frequency encoding with
$6$ bands.  The relightable radiance network has a similar network architecture as
NeuS' color MLP: $4$ hidden layers with $256$ nodes using a ReLU activation.
The final color is outputted after a Sigmoid activation, ensuring that the
output color is within the $(-1, 1)$ range.  Figure \ref{fig:networks}
details network architecture of our method.

%%%%%%%%%%%%%%%%%%%%%%%%%%%%%%%%%%%%%%%%%%%%%%%%%%%%%%%%%%%%%%%%%%%
%\begin{acks}
%\end{acks}
%%%%%%%%%%%%%%%%%%%%%%%%%%%%%%%%%%%%%%%%%%%%%%%%%%%%%%%%%%%%%%%%%%%
%\nocite{*}
\bibliographystyle{ACM-Reference-Format}
\bibliography{src/reference}

%%%%%%%%%%%%%%%%%%%%%%%%%%%%%%%%%%%%%%%%%%%%%%%%%%%%%%%%%%%%%%%%%%%